\documentclass[lettersize,journal]{IEEEtran}
\usepackage{amsmath,amsfonts}
\usepackage{algorithmic}
\usepackage{algorithm}
\usepackage{array}
\usepackage[caption=false,font=footnotesize]{subfig}
\usepackage{textcomp}
\usepackage{stfloats}
\usepackage{balance}
\usepackage{url}
\usepackage{bm}
\usepackage{cite}
\usepackage{lipsum}
\usepackage{pifont}
\usepackage{booktabs}
\usepackage{verbatim}
\usepackage{graphicx}
\usepackage{multirow}
\usepackage[table, dvipsnames]{xcolor}
\usepackage{soul}

\definecolor{verylightgray}{gray}{0.95}
\newcolumntype{g}{>{\columncolor{verylightgray}}c}

\newcommand{\bfr}{\color{OrangeRed} \bf}
\newcommand{\bfb}{\color{RoyalBlue} \bf}

\begin{document}

\title{GLIM: 3D Range-Inertial Localization and Mapping \\ with GPU-Accelerated Scan Matching Factors}


\author{Kenji Koide, Masashi Yokozuka, Shuji Oishi, and Atsuhiko Banno
\thanks{This work was supported in part by JSPS KAKENHI Grant Number 23K16979 and a project commissioned by the New Energy and Industrial Technology Development Organization (NEDO).}
}

\maketitle

\newcommand{\xmark}{\ding{55}}%

\begin{abstract}

This article presents GLIM, a 3D range-inertial localization and mapping framework with GPU-accelerated scan matching factors. The odometry estimation module of GLIM employs a combination of fixed-lag smoothing and keyframe-based point cloud matching that makes it possible to deal with a few seconds of completely degenerated range data while efficiently reducing trajectory estimation drift. It also incorporates multi-camera visual feature constraints in a tightly coupled way to further improve the stability and accuracy. The global trajectory optimization module directly minimizes the registration errors between submaps over the entire map. This approach enables us to accurately constrain the relative pose between submaps with a small overlap. Although both the odometry estimation and global trajectory optimization algorithms require much more computation than existing methods, we show that they can be run in real-time due to the careful design of the registration error evaluation algorithm and the entire system to fully leverage GPU parallel processing.

\end{abstract}


\section{Introduction}
\IEEEPARstart{R}{ange}-based simultaneous localization and mapping (SLAM) has been actively studied due to the emergence of various range sensors, from highly accurate light detection and ranging (LiDAR) sensors to affordable depth cameras. It has become inevitable for wide variety of applications, including surveillance and autonomous driving.

Most modern range-based SLAM systems employ a two-stage estimation approach that first estimates the sensor ego-motion in real-time using local sensor data (i.e., odometry estimation) and then corrects accumulated errors by considering the global consistency of the map and trajectory.

For range-based odometry estimation, a combination of scan-to-model point cloud matching \cite{Zhang2016} and tight coupling of range and IMU constraints has been used in many studies \cite{Ye2019, Qin2020,Bai2022}. A notable range-IMU odometry estimation algorithm, FAST-LIO2 \cite{Xu2022}, which has state-of-the-art accuracy with a limited computation cost, fuses the scan-to-model matching constraint and the IMU constraint on an efficient iterated Kalman filter. However, because the scan-to-model matching and state filtering approaches immediately fix the current sensor state estimate as a new observation arrives, they have difficulty dealing with situations where range data become completely degenerate for a moment. Because it is impossible to determine the current sensor state using only past observations in such situations, existing methods based on causal estimation cannot avoid estimation drift or corruption.

Pose graph optimization is the gold standard for global trajectory optimization owing to its efficiency and stability \cite{Grisetti2010}. Based on relative pose observations between frames obtained using scan matching, it minimizes the sum of relative pose errors to correct estimation drift. In pose graph optimization, each relative pose constraint is modeled as a Gaussian distribution (i.e., mean and covariance matrix).  In practice, however, it is difficult to accurately estimate the covariance matrix of a scan matching result because of the nonlinearity and non-continuity of the scan matching objective function \cite{Landry2019}. Furthermore, the mean point (relative pose) is difficult to be obtained accurately when point clouds have only a small overlap. These difficulties lead to inaccurate modeling of relative pose constraints, resulting in deteriorated estimation accuracy for long trajectories with large loops.

Our aim in this work is to overcome the aforementioned shortcomings in existing range-based SLAM systems by introducing an extremely fast GPU-accelerated registration error evaluation algorithm and replacing the standards in both the odometry estimation and global trajectory optimization algorithms, namely scan-to-model matching and pose graph optimization with relative pose constraints.

We first propose an odometry estimation algorithm that fuses point cloud registration error constraints and IMU constraints for low-drift odometry estimation with a bounded computation cost. In contrast to conventional approaches, the proposed approach continually updates past sensor states in the optimization window (i.e., fixed-lag smoothing) and represents the scan matching target model as a set of selected past frames (i.e., keyframes). It propagates the latest sensor state estimate to past ones to correct estimation drift. This approach makes the odometry estimation extremely robust to a few seconds of completely degenerate range data as well as quick sensor motion. Furthermore, we incorporate multi-camera visual constraints in the odometry estimation factor graph in a tightly coupled way to further improve the stability and accuracy.

We then propose a global trajectory optimization algorithm that directly minimizes registration errors between submaps over the entire map (i.e., global multi-scan registration). Unlike pose graph optimization, this method avoids the Gaussian approximation of relative pose constraints and directly computes point cloud registration errors on the factor graph. It allows the relative pose between point clouds with a very small overlap to be accurately constrained. Furthermore, we incorporate IMU constraints into the global optimization factor graph in a tightly coupled way to better stabilize the global optimization and reduce trajectory estimation errors in four degrees of freedom (DoFs).

The proposed algorithms both require much more computation than that for conventional algorithms. Running them in real-time was thus considered to be infeasible. In this work, by carefully designing the registration error evaluation algorithm and the entire framework to fully leverage GPU parallel processing, we show that these algorithms can run in real-time on a single consumer-grade GPU.

{\bf Contributions: }
The main contributions of this work are as follows:

\begin{itemize}
  \item We extend our GPU-accelerated voxelized GICP registration error factor \cite{Koide2022} with surface-orientation-based correspondence validation and a multi-resolution voxelmap to improve its stability and accuracy in indoor environments. We also propose an efficient linearization mechanism for GPU-based factors to minimize CPU-GPU synchronization overhead and maximize  optimization speed.
  \item We propose a keyframe-based tightly coupled range-IMU odometry estimation algorithm that is extremely robust to quick sensor motion and momentary degeneration of range data. We present a multi-camera visual-range-IMU extension to further improve the estimation accuracy.
  \item We propose a global trajectory optimization algorithm based on global registration error minimization with tightly coupled IMU constraints. We introduce the concept of submap {\it endpoints} to strongly constrain submap poses with a large time interval with IMU constraints.
  \item The proposed framework is carefully designed to be general to the sensor model. We show that it can handle various range-IMU sensors including not only LiDARs but also depth and stereo cameras.
  \item We release the majority of the code of the proposed framework, GLIM, as open source\footnote{\url{https://github.com/koide3/glim}}.
\end{itemize}

\section{Related work}

Modern SLAM frameworks comprise two modules: 1) an odometry estimation module that estimates sensor ego-motion in real-time using local sensor data and 2) a global trajectory optimization module that corrects accumulated estimation errors by considering the global consistency. We briefly review recent trends in both odometry estimation and global trajectory optimization for range-based SLAM and explain the concept of the proposed framework in this section.

\subsection{Odometry Estimation}
\label{sec:related_frontend}

{\bf Feature vs. direct point cloud registration:}
Point cloud registration is a key element for sensor ego-motion estimation. Two types of method for point cloud registration are commonly used for range-based SLAM, namely feature-based approaches, which extract plane and edge points and align point clouds by matching these points, and direct approaches, which avoid feature extraction and use most of the input points to align point clouds.

Feature-based point cloud registration was first proposed for LOAM \cite{Zhang2016}. It extracts plane and edge points based on local smoothness, and then aligns point clouds by minimizing point-to-edge and point-to-plane matching distances. Usually, a process called frame-to-model matching follows the frame-by-frame scan matching to reduce the estimation drift. In the frame-to-model matching, aligned point clouds are accumulated in a single target point cloud (i.e., map model) and the sensor pose is estimated by aligning the current scan with the model point cloud. Many works employed a combination of feature-based registration and frame-to-model matching \cite{Shan2018,liosam2020shan,clins} because of its efficiency. This approach requires careful tuning of feature extraction, and feature extraction algorithms dedicated to specific sensors are often used \cite{Shan2018,Qin2020,liosam2020shan}.

Direct point cloud registration avoids feature extraction and uses most points to perform scan matching \cite{Behley2018,Wang2022a,Chen2022,Wang2022,Xu2022}. Variants of the ICP algorithm with a local surface model (e.g., point-to-normal ICP \cite{Behley2018} and generalized ICP \cite{Segal2009,Chen2022}) are often used owing to their accuracy and robustness. Although the computation cost of direct registration methods is higher than that of feature-based approaches, the former are more robust to sensor and environment changes because they do not require a delicate feature extraction process. The scan-to-model matching approach is often used for direct registration methods as well \cite{Behley2018,Chen2022,Wang2022,Xu2022}.

{\bf Loose vs. tight IMU coupling:} 
With progress in visual-inertial SLAM \cite{Qin2018,Stumberg2018,Campos2021}, IMU fusion has become an important technique for range-based SLAM \cite{liosam2020shan}. The use of an IMU enables the prediction of sensor motion at a frequency of 100 to 1000 Hz, facilitating good initial estimates of the sensor pose and distortion correction of point clouds under quick sensor motion. Furthermore, IMU measurements provide information on the direction of gravity, enabling a reduction of trajectory estimation drift in four DoFs \cite{Qin2018}.

IMU and point cloud measurements can be fused using a loose coupling scheme, which separately considers range-based estimation and IMU-based estimation and fuses the estimation results in the pose space using a method such as the extended Kalman filter \cite{Weiss2011} or factor graph optimization \cite{liosam2020shan,Indelman2013}. Because this approach enables heterogeneous sensor data (e.g., LiDAR, camera, and IMU) to be easily fused, several methods employ an IMU-centric loose coupling approach to robustly estimate the sensor state in extreme environments (e.g., an underground mine) \cite{Zhao2021,Palieri2021,Reinke2022}. Although a loose coupling scheme is computationally efficient and can easily handle heterogeneous sensor data, it is difficult to accurately propagate and fuse the uncertainty for each sensor with a loose coupling scheme, and tight coupling schemes are more accurate and robust than loose coupling schemes \cite{Qin2018,Ye2019}. 

A tight coupling scheme estimates the sensor states by directly minimizing a unified objective function that combines range- and IMU-based constraints. This approach enables the robust estimation of sensor states in environments where sufficient geometric information is not available through range data because the IMU constraints help to constrain the sensor state based on inertial information. Owing to their accuracy and robustness, tightly coupled LiDAR-IMU methods have been widely studied in recent years \cite{Ye2019,Qin2020,Xu2021,Li2021}. 

{\bf Filtering vs. optimization:} 
Tight coupling schemes can be categorized into filtering- and optimization-based approaches. The filtering approach keeps only the latest sensor state and immediately marginalizes old states when a new observation arrives. The optimization approach keeps past frames active and optimizes them via nonlinear optimization. To limit the computation cost, only a subset of past frames (e.g., a number of keyframes or frames in a sliding window) is usually considered for optimization. Although the filtering approach is efficient since it holds and optimizes only the latest sensor state, it quickly accumulates linearization errors. In the context of visual SLAM, it has been shown that the optimization-based approach outperforms the filtering-based approach in terms of accuracy because it re-linearizes past measurements \cite{Strasdat2012, Forster2017}.

Despite the theoretical advantage of the optimization-based approach, in the context of range-based SLAM, many studies have used the filtering approach for IMU fusion \cite{Qin2020,Xu2022}. This was likely performed for two reasons. First, point cloud matching is more computationally expensive than visual feature matching and jointly considering multiple frames in a few seconds of the optimization window in real-time is considered to be infeasible. Second, the widely used frame-to-model matching approach needs to immediately fix the sensor pose estimate to accumulate points into a target model point cloud. Updating the sensor state estimates of past frames does not make a significant difference because the model point cloud is already frozen.

{\bf Continuous-time pose representation:}
LiDARs typically use a laser sweeping mechanism for wide field-of-view measurements that can cause point cloud distortion when the sensor is in motion. To compensate for this distortion, some works have used continuous-time sensor pose representation based on pose interpolation, which enables a sensor pose at any time step to be obtained \cite{clins, Dellenbach_2022, Droeschel_2018, Le_Gentil_2021}. Compared to the discrete-time pose representation, the continuous-time representation allows for obtaining a sensor pose at any time step by nature and thus facilitates compensation of motion distortion on a per-point basis. It also enables incorporating IMU constraints without pre-integration by considering derivatives of the interpolated pose.

We, however, avoid the continuous-time representation and employ the conventional discrete-time representation in this work for two reasons. Firstly, emerging range sensors without a laser sweeping mechanism provide distortion-free point clouds with points acquired simultaneously, rendering continuous-time deskewing unnecessary (e.g., Microsoft Azure Kinect, Intel Realsense, and solid state LiDARs). Secondly, in the context of visual SLAM, it has been reported that the continuous-time representation does not improve estimation accuracy compared to the discrete counterpart, as long as the camera is well-synchronized with the IMU \cite{Cioffi_2022}. We thus consider the continuous-time representation unnecessary when point cloud distortion is not significant and can sufficiently be cancelled by IMU-based motion prediction.

{\bf Proposal:} We propose a tightly coupled range-IMU odometry estimation algorithm based on keyframe-based point cloud matching and fixed-lag smoothing. In contrast to the conventional filtering-based approaches, which only optimize the latest sensor state, the proposed method keeps each frame active for a few seconds and enables dealing with momentary degeneration of range data by propagating the sensor states of successive frames to past frames. To make the framework generic (i.e., suitable for any range sensor), we use a direct point cloud registration approach based on a distribution-to-distribution comparison. Although these design choices lead to an enormous amount of computation, we show that it is feasible to run the algorithm in real-time by fully leveraging a modern GPU.

\subsection{Global Trajectory Optimization}
\label{sec:related_backend}

{\bf Pose graph optimization:} Pose graph optimization constructs a factor graph with (SE3) relative pose constraints and optimizes the sensor poses by minimizing the errors in the pose space. This approach has been widely used for global trajectory optimization in many SLAM systems owing to its simplicity and efficiency \cite{Qin2018,Shan2018,liosam2020shan}. Pose graph optimization models each relative pose constraint as a Gaussian distribution on a SE3 manifold. In practice, however, it is difficult to accurately estimate the covariance matrix of a relative pose given by scan matching. Although a closed-form method is commonly used to estimate the covariance matrix from the Hessian matrix of an ICP scan matching result \cite{Censi2007}, this approach tends to be optimistic because it considers only the last linearized objective function and cannot take into account point correspondence changes \cite{Landry2019}. Although Monte-Carlo-based covariance estimation methods \cite{Iversen2017} are more accurate than the Hessian-based method, they require scan matching to be performed multiple times, resulting in a large computation cost. Landry et al. proposed a learning-based method that takes point cloud descriptors and estimates the covariance matrix based on an offline trained model. While this approach aims to balance estimation accuracy and processing cost, its estimation accuracy can deteriorate in unseen environments. Most existing range-based SLAM frameworks use only a constant covariance matrix \cite{Behley2018}, a simple weighting scheme \cite{Shan2018,liosam2020shan}, or Hessian-based closed-form covariance estimation \cite{Hess2016}. Such inaccurate constraint modeling deteriorates the estimation accuracy of long trajectories with large loops.

For point clouds with only a small overlap, it is difficult to explicitly determine the relative pose between the point clouds via scan matching. Thus, existing frameworks, which rely on pose graph optimization, close loops only when there is sufficient overlap between point clouds, and discard information on point cloud pairs with a small overlap. If we forcibly close loops between such frames, the corrupted scan matching results turn into inaccurate relative constraints and result in deteriorated trajectory estimation \cite{koide_ral2021}.

{\bf Bundle adjustment:} Bundle adjustment (BA) is an approach for simultaneously estimating environment parameters (e.g., 3D landmark positions) and sensor states over frames. It is an essential technique for visual SLAM and is widely used for both odometry estimation \cite{Qin2018} and global optimization \cite{Campos2021}. For range-based SLAM, the BA problem is often formulated as the joint optimization of line and plane environment parameters and sensor poses \cite{Zhou2021,Wisth_2023}. Liu and Zhang pointed out that line and plane parameters can be eliminated from the optimization variables because they can be determined from point coordinates and sensor poses, and formulated line and plane BA as the eigenvalue minimization of accumulated points \cite{liu2020balm}. Huang et al. further re-formulated the plane BA problem in least-squares form for efficient optimization \cite{Huang2021}. Although BA-based methods promise consistent mapping results, they need a good initial guess of the sensor poses for correspondence estimation. In particular, the least squares BA formulation \cite{Huang2021} is sensitive to the initial guess because its surface normal estimation is separated from sensor pose optimization. BA-based methods thus have difficulty in closing loops under a large estimation drift and need to be combined with hierarchical optimization with pose graph \cite{Liu_2023}.

{\bf Global matching cost minimization:} Lu and Milios formulated the 2D range-based mapping problem as the minimization of scan registration errors on a factor graph \cite{Lu1997}. Several works extended this approach to three dimensions by explicitly handling loop closing events to reduce optimization executions \cite{Borrmann2008,Sprickerhof2011}. Because this approach does not explicitly require the relative pose between frames (it just evaluates the registration error between them), it can naturally propagate the point matching uncertainty to the frame uncertainty and accurately constrain the relative pose between frames with only a small overlap. However, this approach is computationally expensive because registration errors need to be re-evaluated over the entire map in each optimization iteration and thus real-time global registration error minimization is considered to be infeasible.

Reijgwart et al. proposed a volumetric mapping framework based on signed distance function submaps \cite{Reijgwart2020}. Their framework creates local submaps in the form of the Euclidean signed distance field (ESDF) and optimizes the submap poses such that the registration errors between them are minimized. Because the distance between a point and a submap surface can be directly obtained with ESDF, efficient registration error computation can be conducted without a costly nearest neighbor search. It is, however, still computationally demanding to compute the global registration error and the optimization is carried out with only a random subset of registration errors with the support of SE3 relative pose constraints.

{\bf IMU constraints in global optimization:} Considering that IMU data provide information on the direction of gravity and suppress odometry estimation drift in yaw and pitch rotation, one can perform global trajectory optimization with four DoF pose graph optimization to keep the global optimization result gravity aligned \cite{Qin2018}. Odometry estimation results, however, may not be perfectly aligned to the gravity direction due to bias estimation errors. In such cases, the four DoF global optimization would result in an inconsistent mapping result.

We can also consider directly inserting IMU constraints into the global optimization factor graph \cite{MurArtal2017a,Schneider2018}. However, keyframes (or submaps) for global optimization are usually created with a larger time interval. If we simply create an IMU factor between keyframes, we cannot strongly constrain their poses because a longer integration time results in a larger uncertainty of the IMU constraint \cite{MurArtal2017a}. Thus, this approach requires the frequent creation of keyframes, which results in redundant computation and a limitation of the factor graph design. 

Usenko et al. proposed a nonlinear factor recovery technique for transferring information accumulated during visual inertial odometry to global optimization \cite{Usenko2020}. In this technique, a set of nonlinear factors are created to approximate a linearized subgraph (Markov bracket) of the odometry estimation factor graph to represent the original distribution with a sparse graph topology. However, because this approach recovers a linearized subgraph, linearization errors are unavoidable.

{\bf Proposal:} Compared to the conventional pose graph optimization approach, the proposed global registration error minimization approach allows the relative pose between submaps with a very small overlap to be accurately constrained, resulting in highly consistent mapping results. As in the proposed odometry estimation algorithm, we leverage GPU parallel processing to perform global matching cost minimization in real-time. Furthermore, we introduce the concept of submap {\it endpoints} to keep the integration time for each IMU factor small and strongly constrain the submap poses with inertial information, which is often unused in existing works.

\section{Proposed System}

\begin{figure*}[tb]
 \centering
 \includegraphics[width=0.95\linewidth]{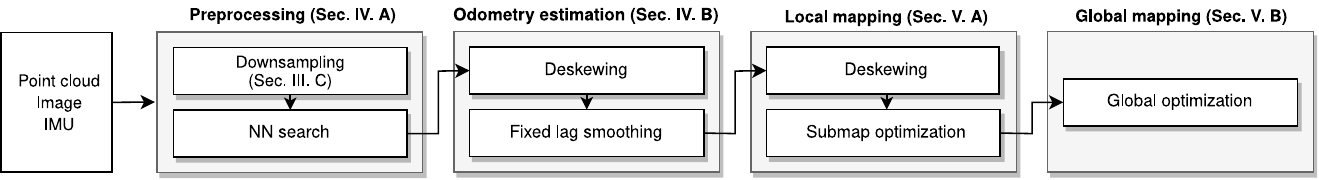}
 \caption{The proposed framework comprises preprocessing module and three estimation modules (odometry estimation, local mapping, and global mapping) that are all based on tight coupling of range and IMU constraints. The input point clouds are fed to the preprocessing module for downsampling and nearest neighbor search. Then, the odometry estimation module performs point cloud deskewing to remove motion distortion, and the sensor motion is estimated through fixed-lag smoothing. The local mapping module performs deskewing again with the odometry estimation result and concatenates several frames into one submap after local map optimization. Finally, the global mapping module takes as input the submaps and optimizes their poses to improve global consistency. Each module runs in an individual thread and the entire system is executed as a single process.}
 \label{fig:system}
\end{figure*}

We give an overview of the proposed framework in Sec. \ref{sec:overview}. We then describe the notation and building blocks for the estimation modules in Sec. \ref{sec:notation} to Sec. \ref{sec:preintegration}. Lastly, we present the proposed odometry estimation and global optimization algorithms in Sec. \ref{sec:frontend} and Sec. \ref{sec:backend}.

\subsection{System Overview}
\label{sec:overview}

Fig. \ref{fig:system} shows an overview of the proposed framework. The mapping system comprises a preprocessing module and three estimation modules, namely odometry estimation, local mapping, and global mapping. All of the estimation modules are based on tight coupling of the range and IMU constraints. Optionally, multi-camera visual information is incorporated in the odometry estimation module to improve the stability and accuracy. We designed the framework to be monolithic for efficiency; each module runs in an individual thread and the entire system is executed as a single process.

\subsection{Notation}
\label{sec:notation}

We define the sensor state ${\bm x}_t$ at time $t$ that will be estimated in the estimation modules as

\begin{align}
{\bm x}_t = [{\bm T}_t, {\bm v}_t, {\bm b}_t],
\end{align}
where ${\bm T}_t = [{\bm R}_t | {\bm t}_t] \in \text{SE}(3)$ is the sensor pose, ${\bm v}_t \in \mathbb{R}^3$ is the velocity, and ${\bm b}_t = [{\bm b}^a_t, {\bm b}^\omega_t] \in \mathbb{R}^6$ is the IMU linear acceleration and angular velocity bias. We estimate the time series of sensor states from point clouds $\mathcal{P}_t$, IMU measurements (linear acceleration ${\bm a}_t$ and angular velocity ${\bm \omega}_t$), and optional camera images $\mathcal{I}_t$. Note that we assume that the transformations between the range sensor, IMU, and camera are known. We transform point clouds into the IMU frame to treat them as if they are in a unified sensor coordinate frame for efficiency and simplicity.

\subsection{Matching Cost Factor}
\label{sec:vgicp}

The matching cost factor constrains two sensor poses (${\bm T}_i$ and ${\bm T}_j$) such that the registration error between the point clouds ($\mathcal{P}_i$ and $\mathcal{P}_j$) is minimized. As the matching cost, we choose voxelized GICP (VGICP) \cite{vgicp}, which is a variant of GICP \cite{Segal2009} with a voxel-based data association. To improve the estimation accuracy in indoor environments, we extend our previous VGICP implementation \cite{vgicp} with surface-orientation-based correspondence validation and a multi-resolution voxelmap.

\begin{figure}[tb]
  \centering
  \includegraphics[width=0.8\linewidth]{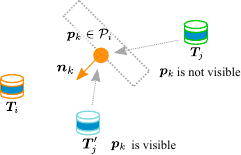}
  \caption{Correspondence validation based on surface orientation. ${\bm p}_k \in \mathcal{P}_i$ is not visible from ${\bm T}_j$, which is on the opposite side of the surface, but is visible from ${\bm T}'_j$, which is on the same side of the surface.}
  \label{fig:surface}
\end{figure}

{\bf Preprocessing:} VGICP models each input point ${\bm p}_k \in \mathcal{P}_i$ as a Gaussian distribution ${\bm p}_k = ({\bm \mu}_k, {\bm C}_k)$, which represents the local shape of the surface around ${\bm p}_k$. The covariance matrix ${\bm C}_k$ is calculated from neighboring points of ${\bm p}_k$ given by a k-nearest-neighbor search. We then create a sparse voxelmap with spatial voxel hashing \cite{Niesner2013} and take the average of the points and their covariances in each voxel. To enlarge the basin of convergence, we create voxelmaps with $L$ resolution levels (e.g., three levels) scaled by a power of two; the voxel size of the $l$-th voxelmap is given by $r^l = r_0 2^{(l - 1)}$, where $r_0$ is the base voxel size. An input point is associated with corresponding voxels in each of the resolution levels to compute the point-to-voxel distance.

Note that comparing a point and the averaged distribution of a voxel is equivalent to comparing the point and all of the points in the voxel in a distribution-to-distribution comparison, as shown in \cite{vgicp}. Because we can obtain a valid distribution even on a distant voxel with only a few points, this approach results in better registration accuracy compared to that for normal distributions transform (NDT) \cite{Bibera}, which needs many points for each voxel to obtain a proper voxel distribution.

\begin{figure*}[tb]
 \centering
 \includegraphics[width=0.8\linewidth]{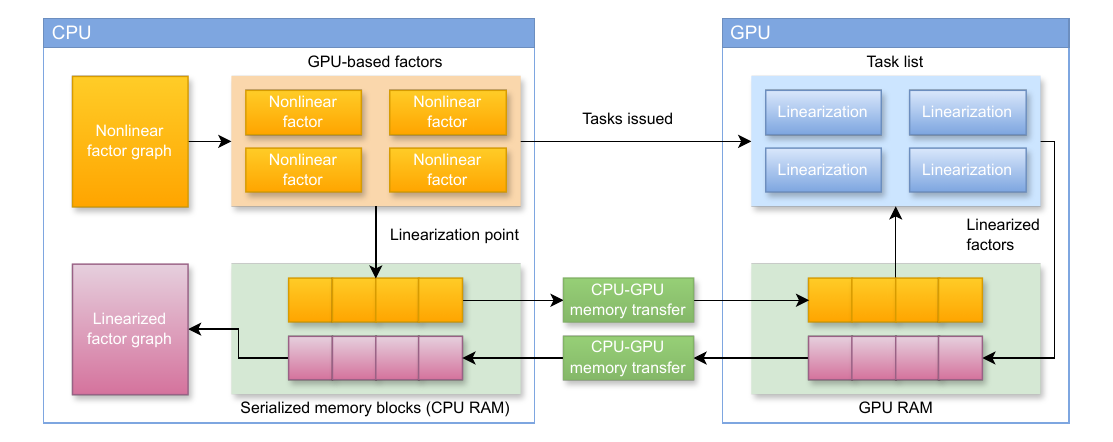}
 \caption{Efficient factor linearization on GPU. The input and output data for linearization are serialized in a memory block and uploaded to and downloaded from the GPU with a single memory transfer. Linearization tasks are issued to the GPU task list and processed in parallel. This mechanism invokes only two CPU-GPU data transfer calls per factor graph linearization task regardless of the number of factors in the factor graph.}
 \label{fig:linearization}
\end{figure*}

{\bf Correspondence search:} In indoor environments, both sides of a thin wall are often observed as the sensor moves. In such cases, a proximity-based correspondence search can wrongly associate points on one side of the wall with points on the other side. To avoid this, we check whether each input point ${\bm p}_k \in \mathcal{P}_i$ is visible from the viewpoint of the other point cloud $\mathcal{P}_j$ based on the surface normal ${\bm n}_k$, as shown in Fig. \ref{fig:surface}. If $ ({\bm p}_k - {\bm T}_i^{-1} {\bm t}_j) \cdot {\bm n}_k > 0$, ${\bm T}_j$ is on the other side of the surface and the point is considered to be not visible from ${\bm T}_j$ and discarded. Otherwise, we consider that ${\bm p}_k$ is visible from ${\bm T}_j$ and calculate the corresponding voxel coordinates ${\bm q}_k = \text{floor}({\bm p}_k / r) \in \mathbb{Z}^3$, where $r$ is the voxel resolution. The corresponding voxels at each of the voxel resolution levels are looked up from the voxelmaps of $\mathcal{P}_j$. This simple correspondence validation works effectively with the odometry estimation based on frame-by-frame comparison proposed in Sec. \ref{sec:frontend}.

{\bf Linearization:} As the objective function to be minimized, we calculate the sum of distribution-to-distribution distances between points ${\bm p_k} = ({\bm \mu}_k, {\bm C}_k)$ and their corresponding voxels $\tilde{{\bm p}}^l_k = (\tilde{\bm \mu}^l_k, \tilde{\bm C}^l_k)$ as follows:

\begin{align}
\label{eq:em}
e^{\text{\it PC}} ({\bm T}_i, {\bm T}_j) = \sum_{l \in [1 \cdots L]} \sum_{{\bm p}_k \in \mathcal{P}_i} e^{\text{\it D2D}} \left( {\bm p}_k, \tilde{\bm p}^l_k, {\bm T}_i^{-1} {\bm T}_j \right), \\
\label{eq:d2d}
e^{\text{\it D2D}}({\bm p}_k, \tilde{\bm p}^l_k, {\bm T}_{ij}) = {\bm d}_k^T \left( \tilde{\bm C}^l_k + {\bm T}_{ij} {\bm C}_k {\bm T}_{ij}^T \right)^{-1} {\bm d}_k,
\end{align}
where ${\bm T}_{ij} = {\bm T}_i^{-1} {\bm T}_j$ is the relative pose between ${\bm T}_i$ and ${\bm T}_j$ and ${\bm d}_k = \tilde{\bm \mu}^l_k - {\bm T}_{ij} {\bm \mu}_k$ is the residual between ${\bm \mu}_k$ and $\tilde{\bm \mu}^l_k$. As in the original GICP, we assume that the displacements of ${\bm T}_i$ and ${\bm T}_j$ are small in each optimization iteration and use ${\bm \Omega}_k = \left( \tilde{\bm C}^l_k + {\bm T}_{ij} {\bm C}_k {\bm T}_{ij}^T \right)^{-1}$ fixed at the linearization point to turn Eq. \ref{eq:d2d} into a weighted least squares form.

Because $e^{\text{\it PC}}$ is in a weighted least squares form, we can derive a linear quadratic factor, which is composed of information matrices ${\bm H}_{ii}, {\bm H}_{ij}$, and ${\bm H}_{jj}$ and information vectors ${\bm b}_i$ and ${\bm b}_j$, from the first-order derivatives:

\begin{align}
{\bm A}_k &= \frac{\partial {\bm d}_k}{\partial {\bm T}_i} = [{\bm R}_{ij} ({\bm \mu}_k)_{\times}, -{\bm R}_{ij}], \\
{\bm B}_k &= \frac{\partial {\bm d}_k}{\partial {\bm T}_j} = [-({\bm T}_{ij} {\bm p}_k )_{\times}, \mathbb{I}_{3\times3}], \\
{\bm H}_{ii} &= \sum_{k}^{N} {\bm A}_k {\bm \Omega}_k {\bm A}_k^T, \quad {\bm H}_{ij} = \sum_{k}^{N} {\bm A}_k {\bm \Omega}_k {\bm B}_k^T, \\
{\bm H}_{jj} &= \sum_{k}^{N} {\bm B}_k {\bm \Omega}_k {\bm B}_k^T, \\
{\bm b}_i &= \sum_{k}^{N} {\bm A}_k {\bm \Omega}_k {\bm d}_k, \quad {\bm b}_j = \sum_{k}^{N} {\bm B}_k {\bm \Omega}_k {\bm d}_k,
\end{align}
where $()_{\times}$ transforms a vector into a skew symmetric matrix (i.e., the hat map).

{\bf Implementation:}
The correspondence search and the linearization can be efficiently computed on a GPU because they are based on simple memory lookup and linear matrix operations and do not involve conditional branches that cause thread divergence. However, simply linearizing factors one-by-one (in a {\it for} loop) would require CPU-GPU synchronization for each factor, which would lead to a large overhead and would significantly affect processing speed.

To avoid the synchronization overhead and fully leverage the power of a GPU, we developed an efficient linearization mechanism, which is shown in Fig. \ref{fig:linearization}. We first list all GPU-based nonlinear factors (matching cost factors in our case) and serialize all data required for linearization (e.g., linearization points) in a single memory block. The serialized input memory block is uploaded to the GPU with a single memory transfer call. Then, the linearization tasks are issued to the GPU task list and processed via several processing streams in parallel. The linearization results are serialized into a memory block on the GPU and sent back to the CPU after CPU-GPU synchronization. This implementation minimizes GPU overhead and maximizes processing speed because it invokes only two CPU-GPU data transfer calls per factor graph linearization task regardless of the number of factors to be linearized.

\subsection{IMU Preintegration Factor}
\label{sec:preintegration}

We use the IMU preintegration technique \cite{Forster2017} to efficiently incorporate IMU constraints into the factor graph. Given an IMU measurement (${\bm a}_t$ and ${\bm \omega}_t$), the sensor state evolves over time as follows:
\begin{align}
\label{eq:imu_evol_R}
{\bm R}_{t + \Delta t} &= {\bm R}_t \exp \left( \left( {\bm \omega}_t - {\bm b}_t^{\omega} - {\bm \eta}_k^{\omega} \right) \Delta t \right), \\
\label{eq:imu_evol_v}
{\bm v}_{t + \Delta t} &= {\bm v}_t + {\bm g} \Delta t + {\bm R}_t \left( {\bm a}_t - {\bm b}_t^a - {\bm \eta}_t^a \right) \Delta t, \\
\label{eq:imu_evol_p}
{\bm t}_{t + \Delta t} &= {\bm t}_t + {\bm v}_t \Delta t + \frac{1}{2} {\bm g} \Delta t^2 + \frac{1}{2} {\bm R}_t \left( {\bm a}_t - {\bm b}_t^a - {\bm \eta}_t^a \right) \Delta t^2,
\end{align}
where ${\bm g}$ is the gravity vector and ${\bm \eta}_t^a$ and ${\bm \eta}_t^{\omega}$ represent white noise in the IMU measurement.

The IMU preintegration factor integrates the system evolution between two time steps $i$ and $j$ to obtain the relative body motion $\Delta {\bm R}_{ij}$, $\Delta {\bm t}_{ij}$, and $\Delta {\bm v}_{ij}$ (see \cite{Forster2017} for a detailed derivation) to constrain the sensor states:

\begin{align}
\begin{split}
\label{eq:preint_R}
e^{\text{\it IMU}} ( {\bm x}_i, {\bm x}_j ) &= \| \log \left( \Delta {\bm R}_{ij}^{T} {\bm R}_i^T {\bm R}_j \right) \|^2 \\
&+ \| \Delta {\bm t}_{ij} - {\bm R}_i^T \left( {\bm t}_j - {\bm t}_i - {\bm v} \Delta t_{ij} - \frac{1}{2} {\bm g} \Delta t_{ij}^2 \right) \|^2 \\
&+ \| \Delta {\bm v}_{ij} - {\bm R}_i^T \left( {\bm v}_j - {\bm v}_i -{\bm g} \Delta t_{ij} \right) \|^2 .
\end{split}
\end{align}

The IMU preintegration factor enables us to keep the factor graph well constrained in environments where geometric features are insufficient and matching cost factors can be degenerated. Furthermore, it provides information on the direction of gravity and reduces the estimation drift in four DoFs \cite{Qin2018}.

\section{Odometry Estimation}
\label{sec:frontend}

In this section, we present the odometry estimation algorithm based on range and IMU data fusion. The key difference compared to existing methods is that the proposed algorithm employs a combination of fixed-lag smoothing and keyframe-based point cloud matching rather than frame-to-model matching and filtering-based estimation. Fixed-lag smoothing continually updates the sensor states when they are in the optimization window. This can be considered to be frame-to-model matching with an active target model. This approach makes the estimation robust to momentary degeneration of range data because it puts sensor state estimates on hold for a few seconds and propagates the latest sensor state estimate to the past ones to correct estimation drift once sufficient geometric constraints on range data become available. Furthermore, unlike filtering-based methods, variables and constraints do not need to be inserted in a strict time order; we can insert constraints for past sensor states as long as the states are still in the optimization window. This enables us to asynchronously run additional data processing (e.g., visual feature tracking) and insert constraints into the factor graph with a slight delay. 

\subsection{Preprocessing}
\label{sec:preprocess}

We first downsample the input point cloud and find $k$ neighboring points for each point that are required for the subsequent point covariance estimation performed after motion distortion compensation. We assume that the neighborhood relationship of points does not change significantly during motion distortion compensation and use the precomputed nearest neighbors for covariance estimation to reduce the processing time. Note that the costly exact nearest neighbor search is only performed in the preprocessing step; all following estimation steps use a voxel-based approximate nearest neighbor search.

\begin{figure}[tb]
 \centering
 \includegraphics[width=1.0\linewidth]{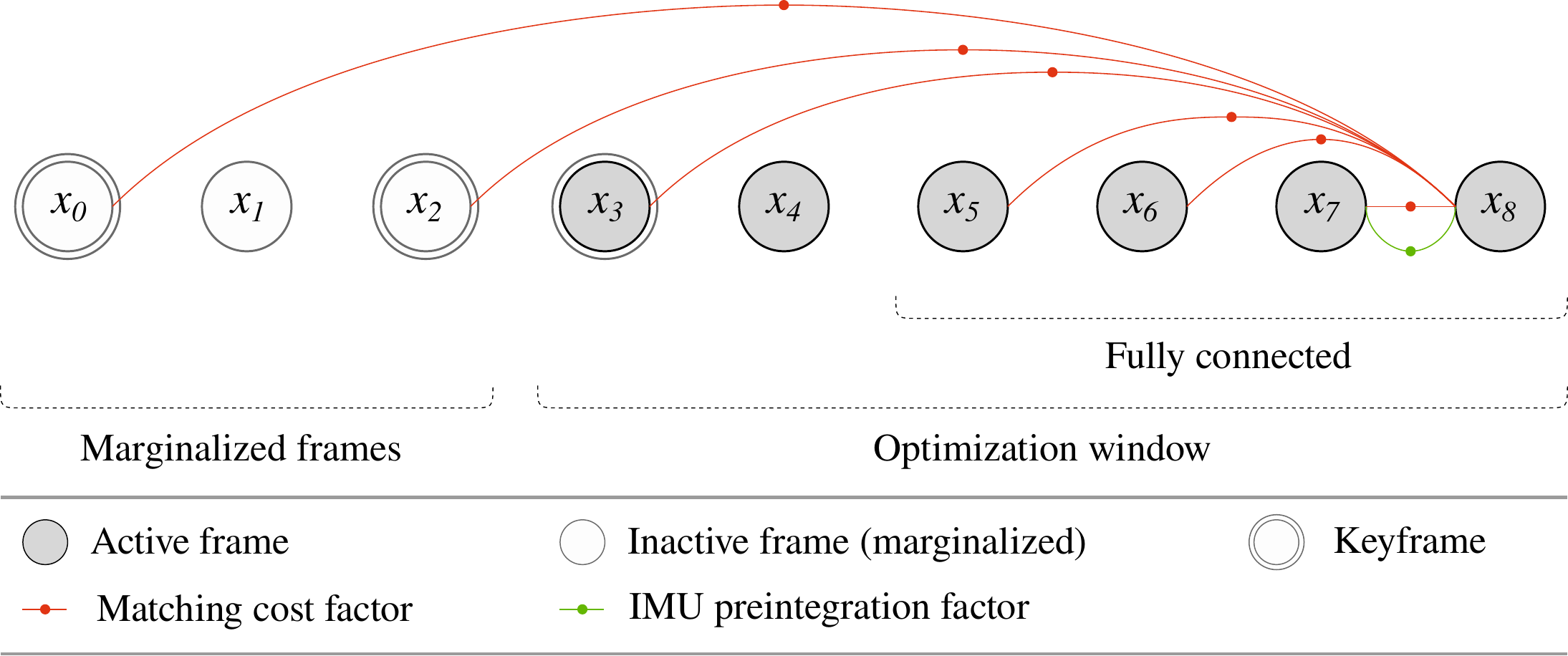}
 \caption{Odometry estimation factor graph. Only factors related to the latest frame (${\bm x}_8$) are visualized. The latest frame is connected with keyframes via matching cost factors to reduce the estimation drift. It is also connected with the last few frames to make the estimation robust to quick sensor motion. IMU factors are created between consecutive frames to keep the factor graph well constrained.}
 \label{fig:frontend_graph}
\end{figure}

\subsection{Tightly Coupled Range-IMU Odometry Estimation}
\label{sec:range_imu_frontend}

We first correct the distortion of the point cloud caused by sensor motion by transforming points into the IMU frame with motion prediction based on IMU dynamics. We then compute the covariance of each point using the precomputed neighboring points.

Given the point clouds and IMU measurements, we construct the factor graph shown in Fig. \ref{fig:frontend_graph}. To limit computation cost and ensure that the odometry estimation algorithm is real-time capable, we use a fixed-lag smoothing approach and marginalize the old frames that move out of the optimization window. We use iSAM2 \cite{Kaess2011} and its efficient Bayes-tree-based variable elimination for fixed-lag smoothing implemented in GTSAM \cite{gtsam}.

Inspired by direct sparse odometry \cite{Engel2018}, we introduce a keyframe mechanism for efficient and low-drift trajectory estimation. Keyframes are a set of frames that are selected such that they are spatially well distributed while having sufficient overlap with the latest frame. We create a matching cost factor between the latest frame and each keyframe to efficiently reduce estimation drift. If a keyframe is already marginalized by the fixed-lag smoother, we consider the keyframe pose as fixed and create a matching cost factor that constrains the latest sensor pose with respect to the fixed keyframe.

To manage keyframes, we define an overlap rate between two point clouds $\mathcal{P}_i$ and $\mathcal{P}_j$ as the fraction of points in $\mathcal{P}_i$ that fall within a voxel of $\mathcal{P}_j$ \cite{koide_ral2021}. Whenever a new point cloud frame arrives, we evaluate the overlap rate between that frame and the union of all keyframes, and if the overlap is smaller than a threshold (e.g., 90\%), we insert that frame into the keyframe list. Similar to the keyframe marginalization strategy in \cite{Engel2018}, we remove redundant keyframes using the following strategy:

\begin{enumerate}
  \item We remove keyframes that overlap the latest keyframe by less than a certain threshold (e.g., 5\%).
  \item If more than $N^{\text{\it odom}}$ (e.g., 20) frames exist in the keyframe list, we remove the keyframe that minimizes the following score:
    \begin{align}
      s(i) = o(i, N^{\text{\it odom}}) \sum_{j \in [1, N^{\text{\it odom}}-1] \backslash \{i\}} \left( 1 - o(i, j) \right),
    \end{align}
    where $o(i, j)$ is the overlap rate between the $i$-th and $j$-th keyframes. The score function is heuristically designed to keep keyframes spatially well distributed while leaving more keyframes close to the latest one.
\end{enumerate}

In addition to the keyframes, we create matching cost factors between the latest frame and the last $N^{\text{\it pre}}$ frames (e.g., last three frames) to make the odometry estimation robust to quick sensor motion. We also create an IMU preintegration factor between consecutive frames for robustness in featureless environments. 

\begin{figure}[tb]
  \centering
  \subfloat[Active frames]{\includegraphics[width=1.0\linewidth]{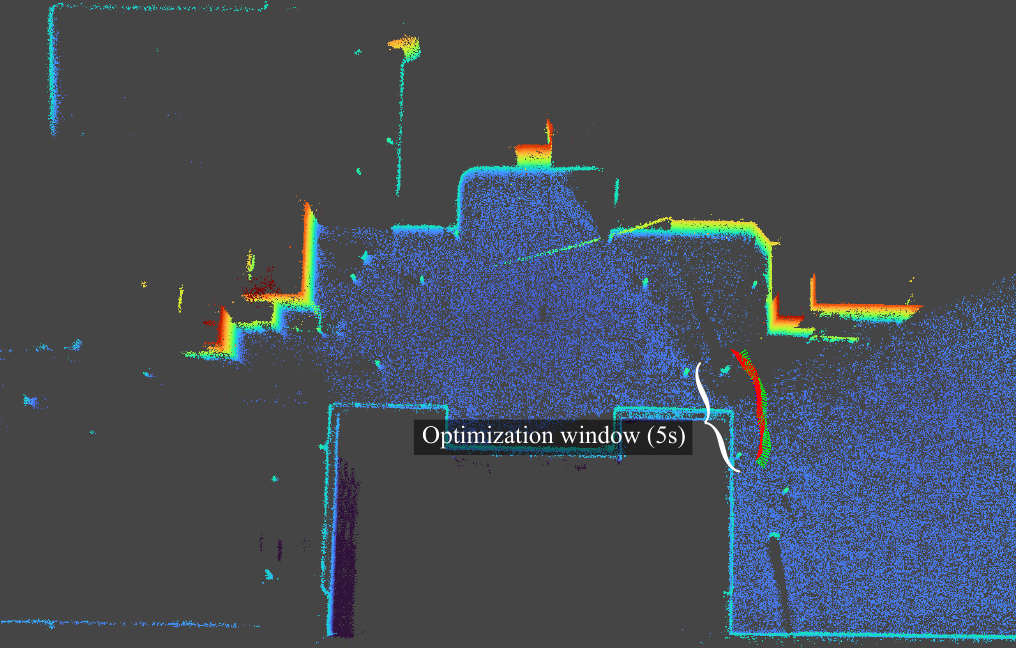}} \ 
  \subfloat[Keyframes]{\includegraphics[width=1.0\linewidth]{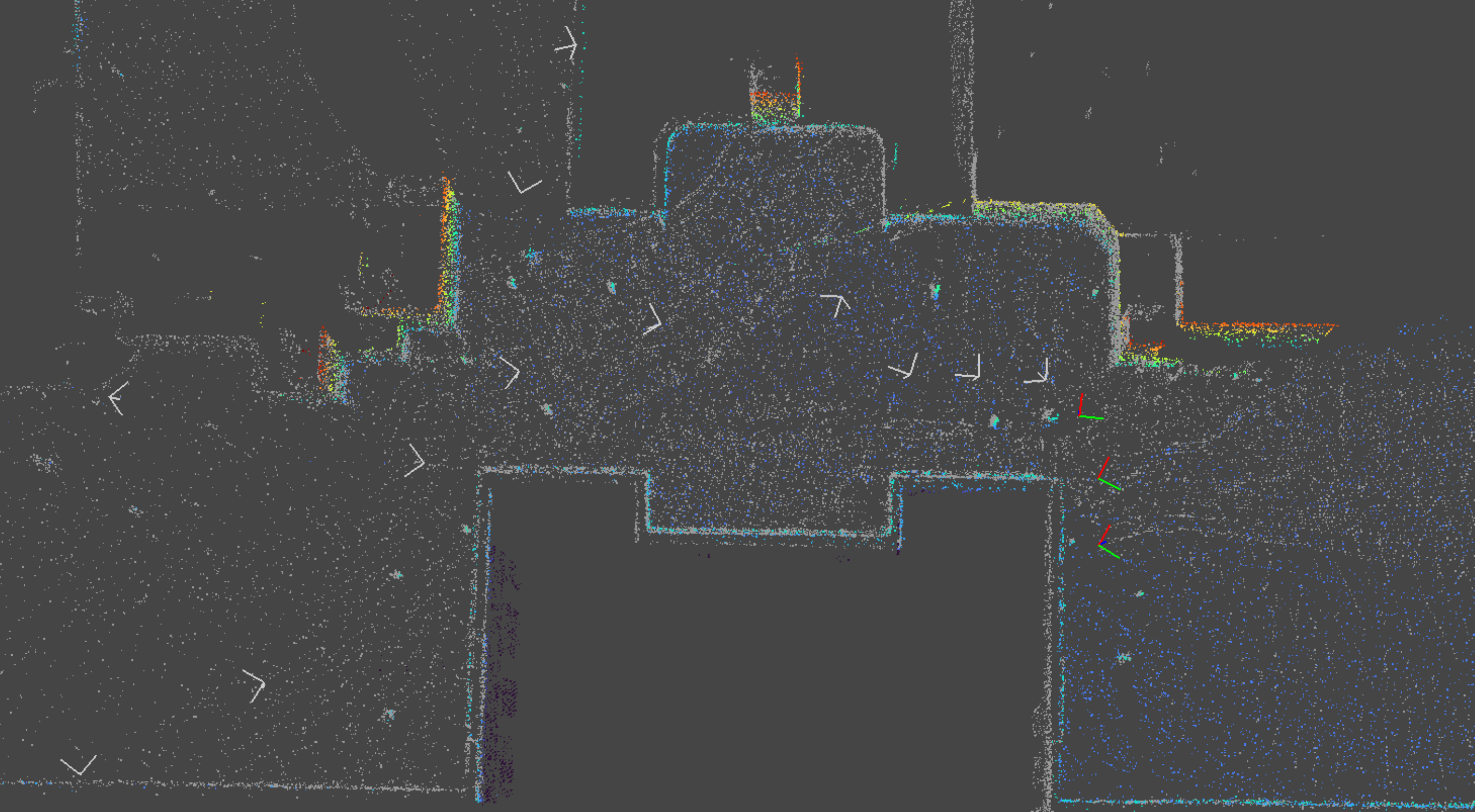}} \ 
  \subfloat[Matching cost factors]{\includegraphics[width=1.0\linewidth]{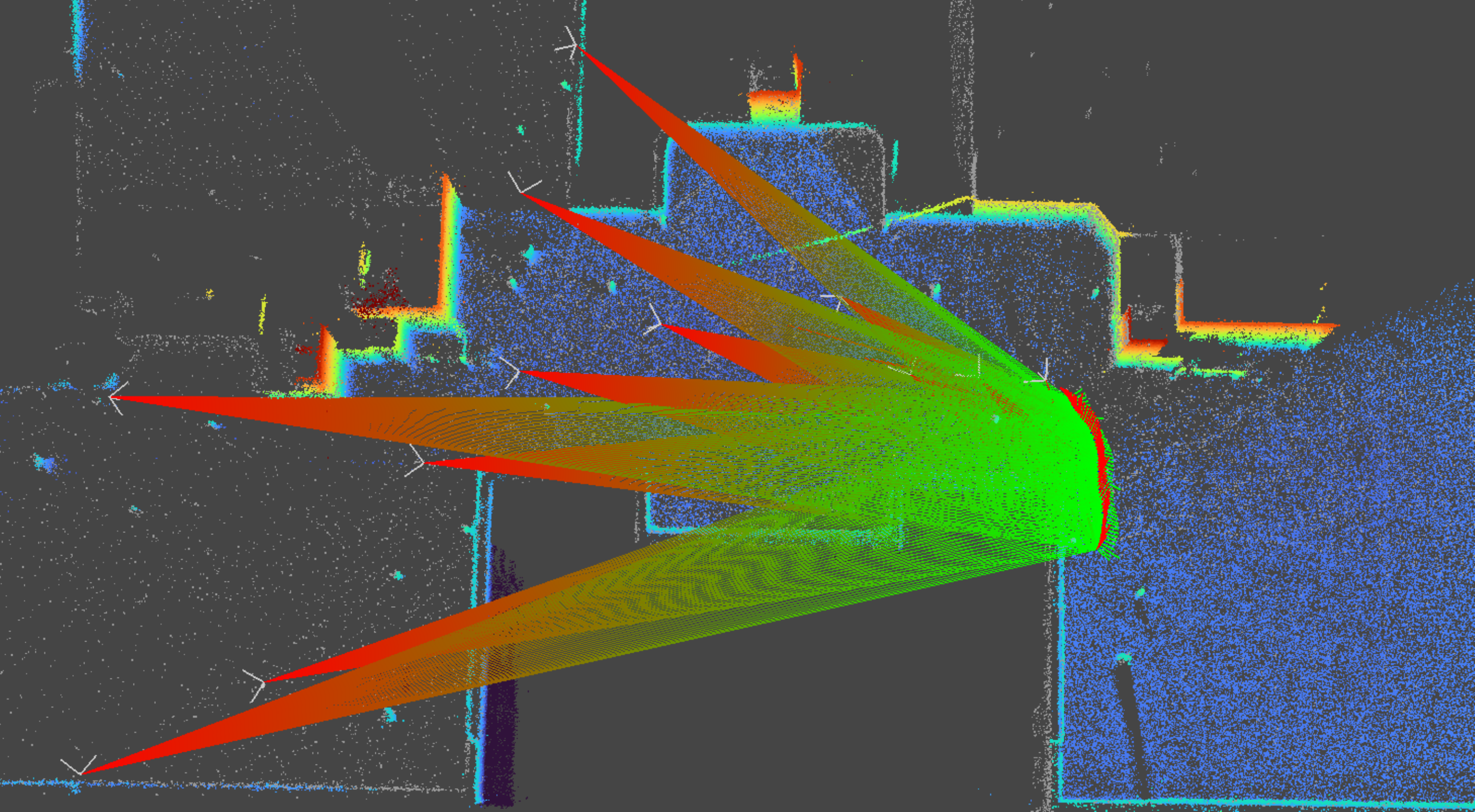}}
  \caption{Example of created keyframes and matching cost factors. Coordinate frames and points shown in color represent active frames in the optimization window (5 s) and those shown in gray are fixed. The matching cost factors are created between the active frames and the keyframes that are selected to be spatially well distributed to efficiently reduce estimation drift.}
  \label{fig:keyframe}
\end{figure}

In summary, the objective function of the odometry estimation algorithm is given by:
\begin{align}
\begin{split}
g^{\text{\it LIO}} (\mathcal{X}^a) 
&= \sum_{{\bm x}_i \in \mathcal{X}^a} \sum_{{\bm x}_j \in \mathcal{X}^p_i \cup \mathcal{X}^k} e^{\text{\it PC}} \left( \mathcal{P}_i, \mathcal{P}_j, {\bm T}_i, {\bm T}_j \right) \\
&+ \sum_{{\bm x}_i \in \mathcal{X}^a} e^{\text{\it IMU}} \left( {\bm x}_{i - 1}, {\bm x}_i \right)
+ e^{\text{\it MG}} \left( \mathcal{X}^a \right),
\end{split}
\end{align}
where $\mathcal{X}^a$ is the set of active frames in the optimization window, $\mathcal{X}^p_i = [ {\bm x}_{i - N^{\text{\it pre}}}, \cdots, {\bm x}_{i - 1} ]$ is the preceding frames of ${\bm x}_i$, $\mathcal{X}^k$ is the keyframes, and $e^{\text{MG}}$ is the error term to compensate for marginalized variables and factors.

Fig. \ref{fig:keyframe} shows examples of active frames, keyframes, and matching cost factors. Coordinate frames and points shown in color represent active frames in the optimization window of the fixed-lag smoother (5 s) and point clouds. Ones shown in gray are the fixed (marginalized) keyframes and point clouds. Fig. \ref{fig:keyframe} (a) shows active frames subject to optimization and Fig. \ref{fig:keyframe} (b) shows keyframes. We can see that the keyframes are selected such that they are spatially well distributed to efficiently reduce estimation drift. The first few keyframes are still in the optimization window and are considered to be active. Fig. \ref{fig:keyframe} (c) shows matching cost factors densely created between the active frames and the keyframes. Although it is difficult to visually confirm, there are also matching cost factors between active frames, as in Fig. \ref{fig:frontend_graph}, that make the estimation robust to quick sensor motion. It is worth emphasizing that we re-evaluate each matching cost factor (each line in Fig. \ref{fig:keyframe} (c)) in each optimization iteration and minimize the sum of registration errors over the densely connected factor graph in real-time.

\subsection{Tightly Coupled Multi-Camera Visual Constraints}
\label{sec:multicam}

We designed the proposed framework to be modular so that it can accept additional constraints to further improve the estimation stability and accuracy. Here, we present a multi-camera extension for tightly coupled visual-range-IMU odometry estimation. 

\begin{figure}[tb]
 \centering
 \includegraphics[width=0.7\linewidth]{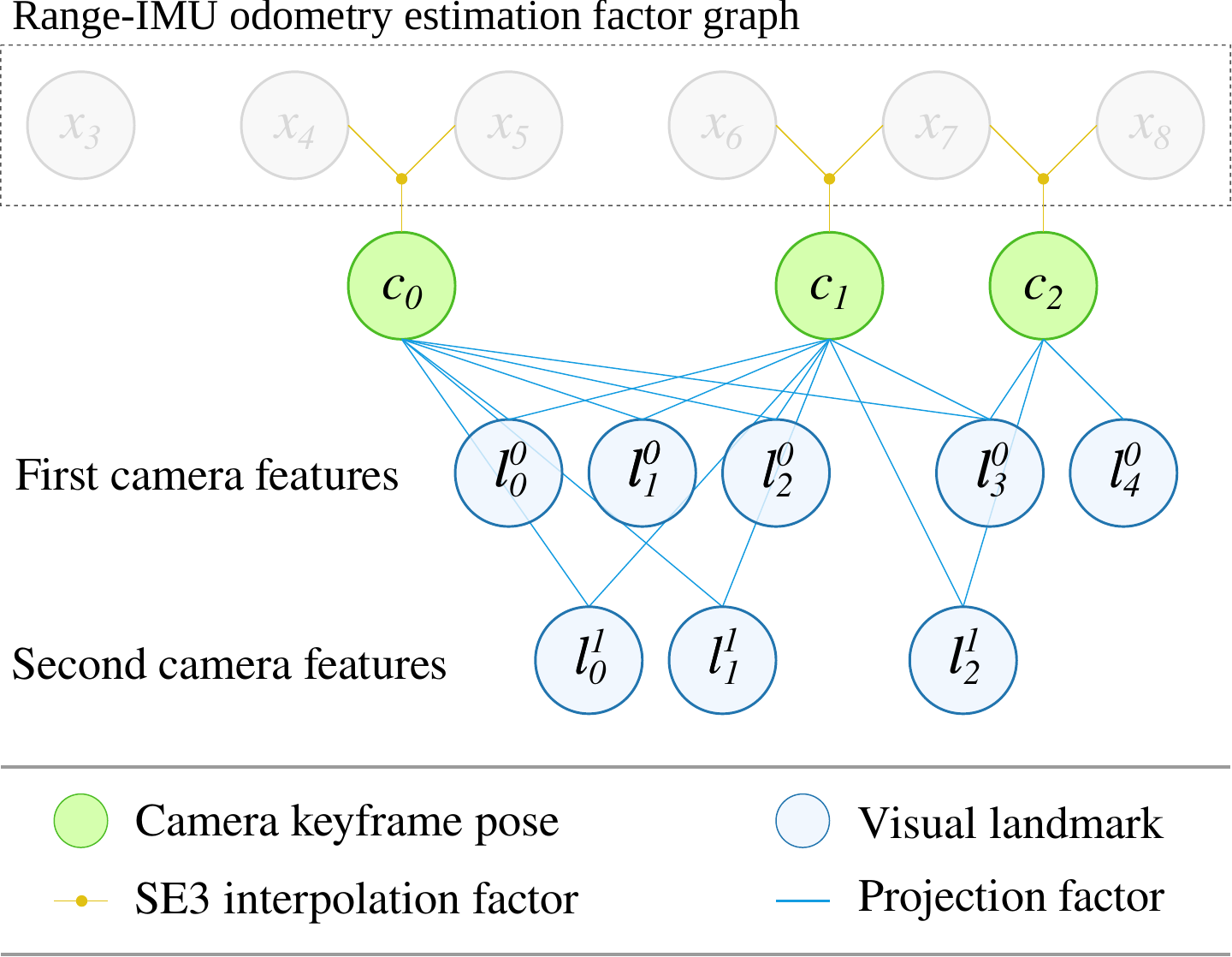}
 \caption{Multi-camera extension for odometry estimation factor graph. Camera keyframe poses are introduced as interpolation of IMU poses and constrained by visual feature projection factors.}
 \label{fig:frontend_graph_multicam}
\end{figure}

Fig. \ref{fig:frontend_graph_multicam} shows the odometry estimation factor graph extended with multi-camera visual constraints. We assume that the cameras are hardware-synchronized and take images at the same moment. For each camera, we detect and track 2D visual feature points using curvature extreme tracking \cite{Yokozuka2019}. Whenever the range sensor measures a certain amount of displacement (e.g., 0.25 m or 15\textdegree) or the time since the last visual keyframe creation exceeds a threshold (e.g., 3 s), we create a visual keyframe ${\bm C}_t \in \text{SE}(3)$ that represents the IMU pose at the moment when an image set was taken. To constrain ${\bm C}_t$, we interpolate the closest two IMU poses ${\bm T}_L = [{\bm R}_L | {\bm t}_L]$ and ${\bm T}_R = [{\bm R}_R | {\bm t}_R]$ that cover the time point of ${\bm C}_t$ using spherical linear interpolation (SLERP):
\begin{align}
{\bm t}^c_t &= (1 - \alpha) {\bm t}_{L} + \alpha {\bm t}_{R}, \\
{\bm R}^c_t &= \text{SLERP}({\bm R}_L, {\bm R}_R, \alpha), \\
\alpha &= \frac{t^C - t^L}{t^R - t^L}, \\
e^{\text{\it VK}} ( {\bm C}_t ) &= \| \log \left( {\bm C}_t^{-1} [ {\bm R}^c_t | {\bm t}_t^c ] \right) \|^2,
\end{align}
where $t^L < t^C < t^R$ are the time points of ${\bm T}_L$, ${\bm C}_t$, and ${\bm T}_R$, respectively.

Everytime a visual keyframe is created, we extract visual features that appear in both the previous and current visual keyframes. We then filter out outlier visual features via a two-stage feature validation. It first removes outliers by using RANSAC-based essential matrix estimation and then triangulates 3D feature positions based on the predicted sensor poses given by LiDAR-IMU fusion. Visual features with reprojection errors larger than a threshold are then eliminated. For each visual feature that passes the validation, we create a projection factor to constrain the corresponding landmark position and visual keyframe pose.

Given a 2D visual feature point $^i{\bm f}_t^j \in \mathbb{R}^2$ tracked by the $i$-th camera at time $t$ with a feature ID $j$, we create a projection factor with Huber's robust kernel defined as:
\begin{align}
e^{\text{\it VF}} ( {\bm C}_t, {\bm l}_i^j ) = \| {^i{\bm f}^j_t} - \rho \left( \pi \left( {\bm C}_t {\bm T}^I_{C_i} {\bm l}^j_i \right) \right) \|^2 ,
\end{align}
where ${\bm l}_i^j \in \mathbb{R}^3$ is the $j$-th visual landmark position associated with the $i$-th camera, ${\bm T}^I_{C_i}$ is the constant transformation between the IMU frame and the optical frame of the $i$-th camera, $\pi$ is the projection function, and $\rho$ is the robust kernel.

The objective function for odometry estimation extended with visual constraints is defined as:
\begin{align}
\begin{split}
g^{\text{\it LVIO}} (\mathcal{X}^a, \mathcal{C}^a, \mathcal{L}^a) &= g^{\text{\it LIO}} (\mathcal{X}^a) + \sum_{{\bm C}_t \in \mathcal{C}^a} e^{\text{\it VK}} ( {\bm C}_t ) \\
& + \sum_{i = 1}^{N^{\text{\it CAM}}} \sum_{{\bm C}_t \in \mathcal{C}^a} \sum_{{\bm l}_i^j \in \mathcal{L}_t^i} e^{\text{\it VF}} ( {\bm C}_t, {\bm l}_i^j ),
\end{split}
\end{align}
where $\mathcal{C}^a$ and $\mathcal{L}^a$ are respectively the sets of visual keyframe poses and visual landmarks in the optimization window, and $\mathcal{L}_t^i$ is a set of visual landmarks that have an observation on $i$-th camera image at time $t$. As shown in Fig. \ref{fig:frontend_graph_multicam}, all of the variables are jointly optimized on a single factor graph by considering multi-sensor constraints (i.e., tightly coupled visual-range-IMU odometry estimation).

\section{Global Trajectory Optimization}
\label{sec:backend}

\subsection{Local Mapping}
\label{sec:local_mapping}

\begin{figure}[tb]
  \centering
  \includegraphics[width=1.0\linewidth]{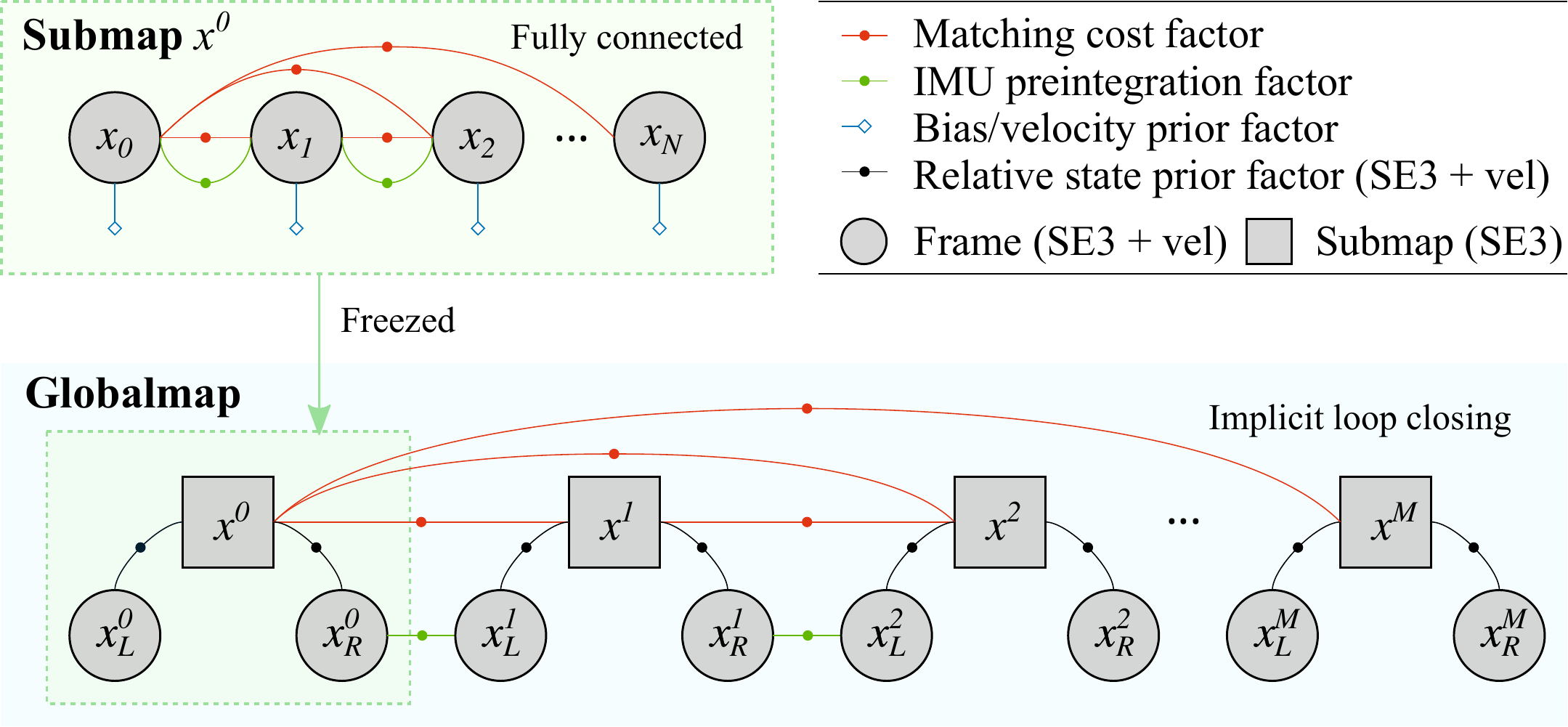}
  \caption{Factor graph for global optimization. The local mapping module refines estimation results and merges several local frames into one submap using an all-to-all registration strategy. The global mapping module optimizes the submap poses such that the global registration error is minimized over the entire map. Both modules take advantage of IMU factors to stabilize the estimation in severe featureless environments and reduce estimation drift.}
  \label{fig:backend_graph}
\end{figure}

Once a frame is marginalized from the odometry estimation graph, it is fed to the local mapping module as an initial estimate of the sensor state. The local mapping module merges several local frames into one submap to reduce the number of optimized variables in the global mapping module.

We first perform deskewing and covariance estimation again with the marginalized state, which is expected to be improved from the initial prediction made at the beginning of the odometry estimation. We then evaluate the overlap rate between that frame and the latest frame in the submap, and if the overlap rate is smaller than a threshold (e.g., 90\%), we insert that frame into the submap factor graph.

As shown in Fig. \ref{fig:backend_graph}, we create a matching cost factor for each combination of frames in the submap (i.e., all-to-all registration). We also add an IMU preintegration factor between consecutive frames and add a prior factor for the velocity and bias of each frame, based on the marginalized state, to better stabilize submap optimization.

The objective function for the local mapping is thus defined as:
\begin{align}
\begin{split}
g^{\text{\it LM}} (\mathcal{X}^s) & = \sum_{ ({\bm x}_i, {\bm x}_j) \in {}_2C_{\mathcal{X}^s} } e^{\text{\it PC}} \left({\bm T}_i, {\bm T}_j \right) \\
& + \sum_{{\bm x}_i \in \mathcal{X}^s} e^{\text{\it IMU}} ( {\bm x}_{i-1}, {\bm x}_i )
+ \sum_{{\bm x}_i \in \mathcal{X}^s} e^{\text{\it BV}} ( {\bm x}_i ),
\end{split}
\end{align}
where $\mathcal{X}^s$ is the set of frames in the submap, ${}_2C_{\mathcal{X}^s}$ is all combinations of the frames, $e^{\text{\it BV}} ({\bm x}_i) = \| {\bm b}_i - {\bm b}_i' \|^2 + \| {\bm v}_i - {\bm v}_i' \|^2$ is a prior factor to keep the velocity and IMU bias estimates close to those of the marginalized state (${\bm b}_i'$ and ${\bm v}_i'$).

Once the number of frames in the submap becomes equal to $N^{\text{\it sub}}$ (e.g., 15), or the overlap between the first and last frames becomes smaller than a threshold (e.g., 5 \%), we perform factor graph optimization using the Levenberg-Marquardt optimizer \cite{Levenberg1944} and merge the frames into one submap based on the optimization result. In contrast to the odometry estimation, we perform batch optimization for local mapping. This enables us to correct estimation drift caused by a longer period of range data degeneration that is difficult to handle through the online optimization with a bounded optimization window used in the odometry estimation.

\subsection{Global Mapping}
\label{sec:global}

The global mapping module optimizes the submap poses such that the registration errors between them are minimized over the entire map. It creates a matching cost factor between each submap pair with an overlap rate that exceeds a small threshold (e.g., 5\%). This results in an extremely dense factor graph. Because each submap is aligned with not only adjacent submaps on the graph but also each revisited submap, trajectory loops are implicitly closed.

Submaps are created with a larger time interval. If we simply create an IMU factor between submap states, its uncertainty becomes very large and it cannot strongly constrain the relative pose between submaps \cite{Stumberg2018,MurArtal2017a}. Furthermore, we also lose information on the velocity and IMU bias estimated by the odometry estimation module. To address these problems, we introduce two states called {\it endpoints} (${\bm x}^i_L$ and ${\bm x}^i_R$) for each submap ${\bm x}^i$; they hold the states of the first and last frames in the submap with respect to the submap pose.

Given an estimate of sensor states $[{\bm x}_1, \cdots, {\bm x}_{N^{\text{\it sub}}}]$ in a submap ${\bm x}^i$, we define the submap origin ${\bm T}^i = [{\bm R}^i | {\bm t}^i]$ as the pose of the sensor at the center ${\bm T}_{N^{\text{\it sub}}/2}$. Then, the sensor state ${\bm x}_t$ relative to the submap origin is given as:
\begin{align}
\label{eq:relative_T}
{\bm T}'_t &= \left( {\bm T}^i \right)^{-1} {\bm T}_t, \\
\label{eq:relative_v}
{\bm v}'_t &= \left( {\bm R}^i \right)^{-1} {\bm v}_t, \\
\label{eq:relative_b}
{\bm b}'_t &= {\bm b}_t.
\end{align}

We consider ${\bm T}'_t$, ${\bm v}'_t$, and ${\bm b}'_t$ to be fixed and create relative state factors between the submap state ${\bm x}^i$ and endpoints ${\bm x}^i_L$ and ${\bm x}^i_R$  respectively from the first and last frames in the submap (${\bm x}_1$ and ${\bm x}_{N^{\text{\it sub}}}$) such that they satisfy the relative state relationships:
\begin{align}
\begin{split}
e^{\text{\it EP}}_L ( {\bm T}^i, {\bm x}^i_L ) &= \| \log \left( {\bm T}^i {\bm T}_1' \left( {\bm T}^i_L \right)^{-1} \right)  \|^2 \\
 & + \| {\bm v}_1' - ({\bm R}^i)^T {\bm v}^i_L \|^2 + \| {\bm b}_1' - {\bm b}^i_L \|^2 ,
\end{split} \\
\begin{split}
e^{\text{\it EP}}_R ( {\bm T}^i, {\bm x}^i_R ) &= \| \log \left( {\bm T}^i {\bm T}_{N^{\text{\it sub}}}' \left( {\bm T}^i_R \right)^{-1} \right)  \|^2 \\
 & + \| {\bm v}_{N^{\text{\it sub}}}' - ({\bm R}^i)^T {\bm v}^i_R \|^2 + \| {\bm b}_{N^{\text{\it sub}}}' - {\bm b}^i_R \|^2 .
\end{split}
\end{align}

We then create an IMU factor between ${\bm x}^i_R$ and ${\bm x}^{i+1}_L$. In this way, an IMU factor covers only a small time interval (range data scan interval) and can strongly constrain the submap poses while avoiding the loss of the velocity and bias information estimated by the local mapping module.

Finally, the objective function for the global trajectory optimization is defined as:
\begin{align}
\begin{split}
g^{\text{\it GM}} ( \mathcal{X}^g ) 
&= \sum_{ ({\bm T}^i, {\bm T}^j) \in \mathcal{X}^o } e^{\text{\it PC}} ( {\bm T}^i, {\bm T}^j ) \\
&+ \sum_{ {\bm T}^i \in \mathcal{X}^g } e^{\text{\it IMU}} ( {\bm x}^{i-1}_R, {\bm x}^i_L ) \\
&+ \sum_{ {\bm T}^i \in \mathcal{X}^g } \left( e^{\text{\it EP}}_L ({\bm T}^i, {\bm x}^i_L) + e^{\text{\it EP}}_R ({\bm T}^i, {\bm x}^i_R) \right),
\end{split}
\end{align}
where $\mathcal{X}^g$ is the set of all submap poses, and $\mathcal{X}^o$ is the pairs of submaps with an overlap.

Whenever a new submap is inserted, the factor graph is incrementally optimized using the iSAM2 optimizer \cite{Kaess2011}.

\section{Experiments}

In this section, we first show the robustness of the proposed odometry estimation algorithm to momentary degeneration of range data in simulated and real environments (Sec. \ref{sec:exp_degenerated}). We then demonstrate that the proposed framework can be applied to various range-IMU sensors (Sec. \ref{sec:exp_various}). Finally, we compare the accuracy of the proposed framework and those of state-of-the-art methods and present ablation studies on the Multi-Camera Newer College Dataset \cite{zhang2021multicamera} (Sec. \ref{sec:exp_newer}) and the NTU VIRAL Dataset \cite{Nguyen_2021} (Sec. \ref{sec:exp_viral})\footnote{The datasets and the estimation results are available online: \url{https://staff.aist.go.jp/k.koide/projects/glimsupp}}.

\subsection{Robustness to Degenerated Range Data}
\label{sec:exp_degenerated}

\begin{figure}[tb]
  \centering
  \includegraphics[width=0.6\linewidth]{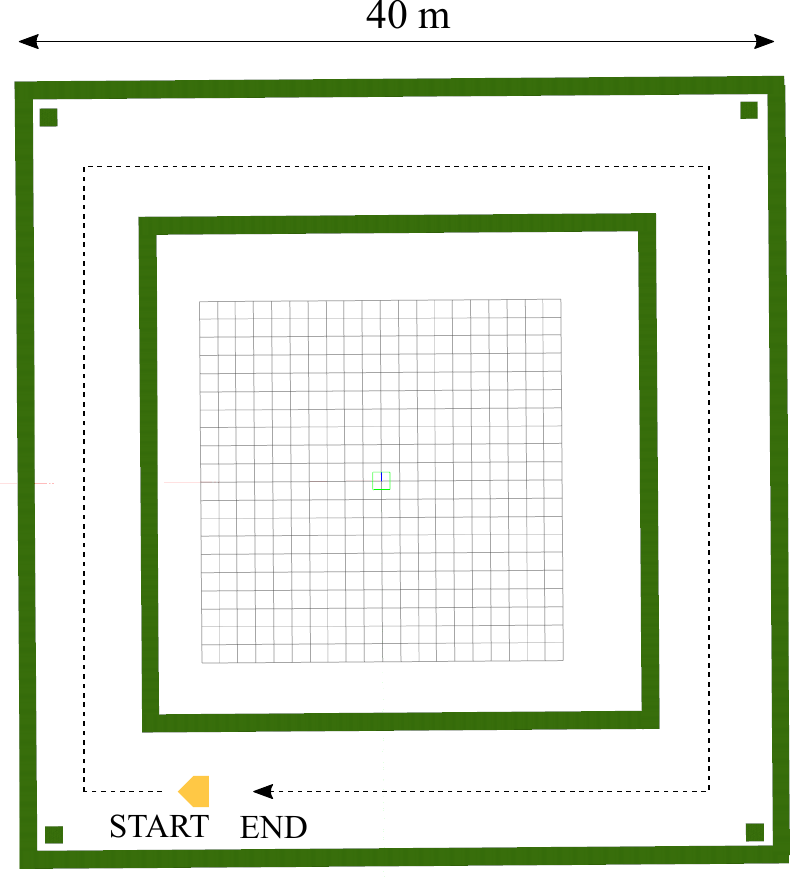}
  \caption{Simulation environment. The environment is 40 m in width. The LiDAR observation range is limited to 15 m. LiDAR data become completely degenerated in the middle of the corridor.}
  \label{fig:simenv}
\end{figure}

\begin{figure}[tb]
  \centering
  \subfloat[LIO-SAM \cite{liosam2020shan}]{\includegraphics[width=0.32\linewidth]{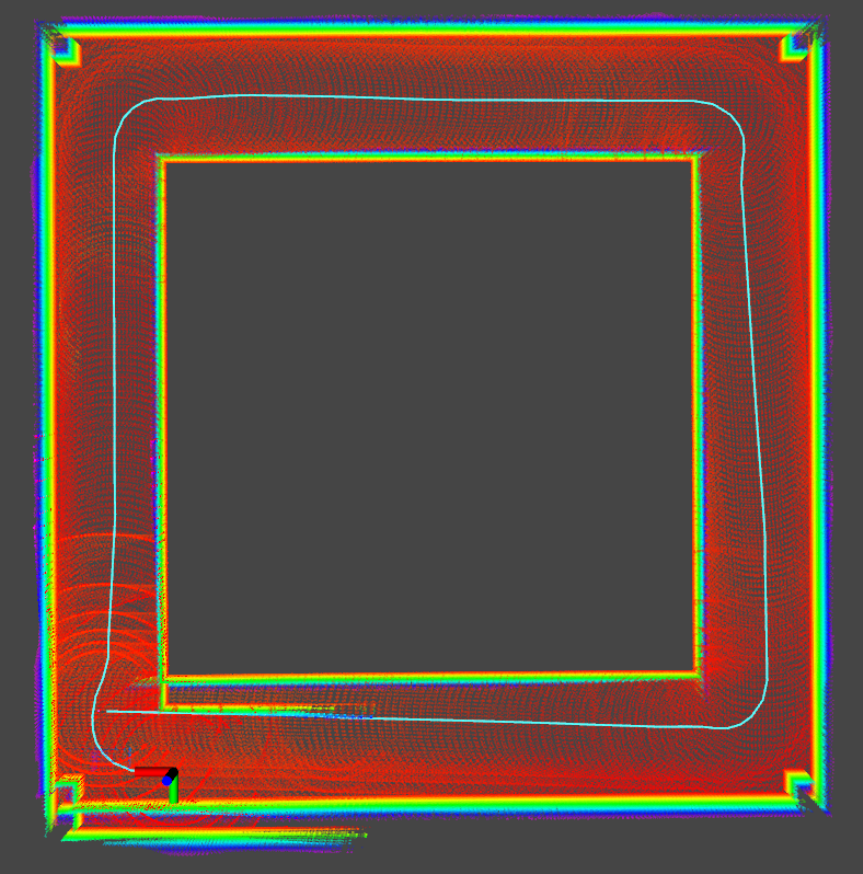}} \ 
  \subfloat[FAST-LIO2 \cite{Xu2022}]{\includegraphics[width=0.32\linewidth]{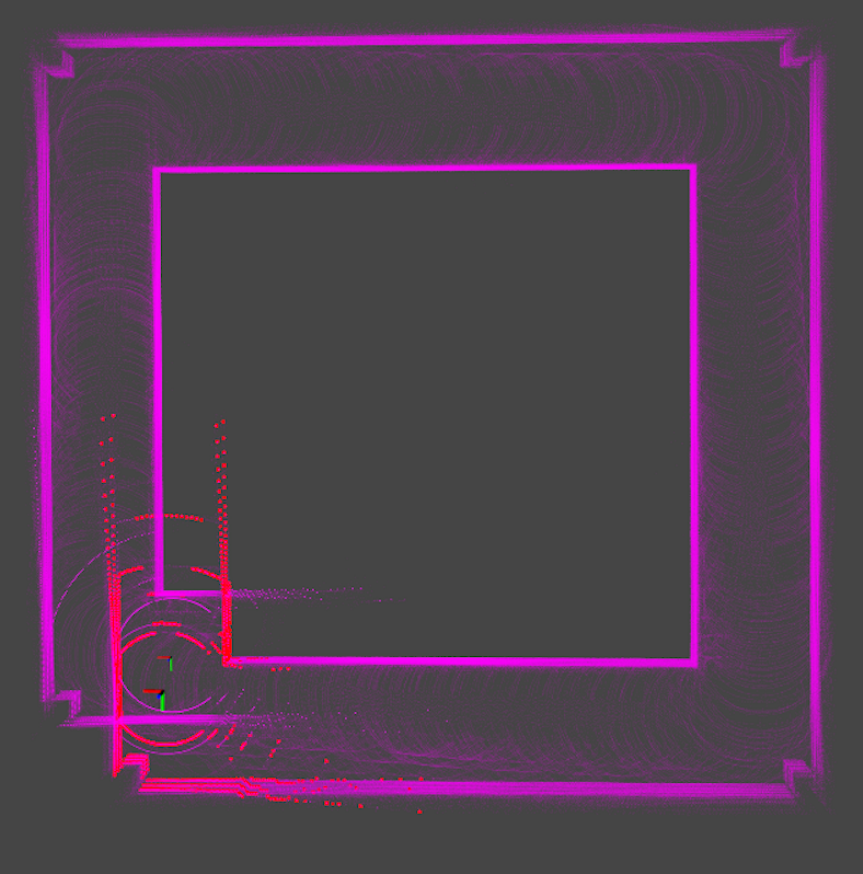}} \ 
  \subfloat[GLIM]{\includegraphics[width=0.32\linewidth]{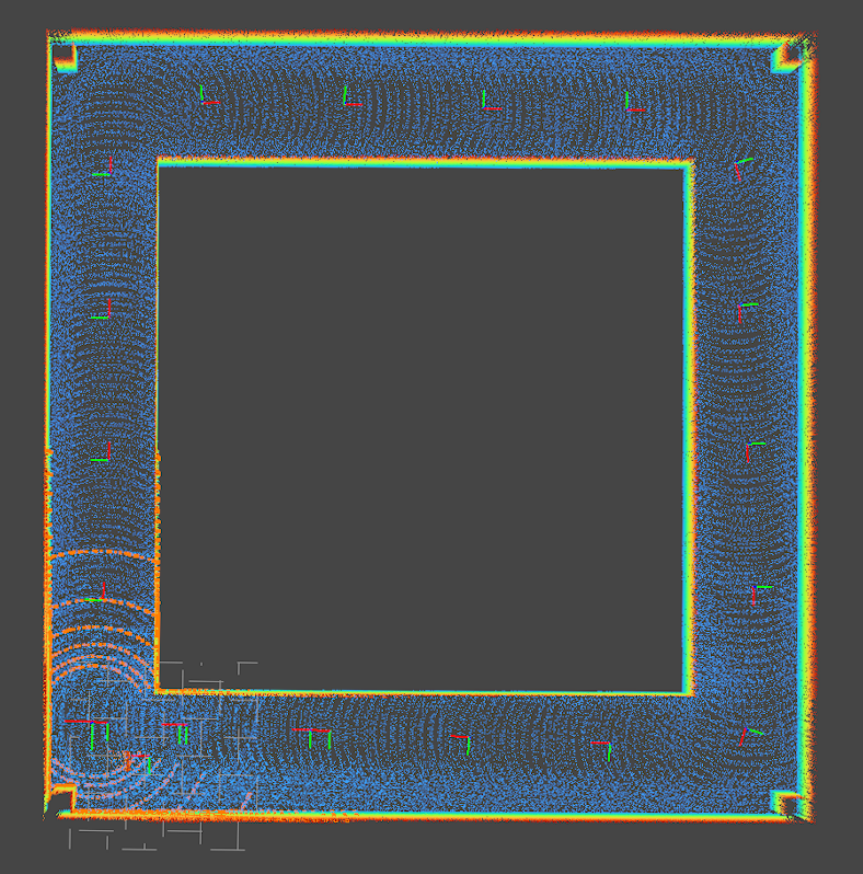}} \ 
  \caption{Estimation results for lowest IMU noise level ($1.0 \times 10^{-3}$ [m/$\text{s}^2$] and [\textdegree / s]). Existing methods based on frame-to-model matching are sensitive to degenerated LiDAR data whereas the proposed method produced a good trajectory estimation result. }
  \label{fig:deg_trajs}
\end{figure}

{\bf Evaluation in simulation: }
To show that the proposed method is robust to momentary range data degeneration, we conducted an evaluation in the simulation environment shown in Fig. \ref{fig:simenv}. We generated point clouds with a LiDAR model (Velodyne VLP-16) while moving it along the corridor, and synthesized IMU data using OpenVINS \cite{Geneva2020}. We generated five LiDAR-IMU sequences while changing the IMU linear acceleration and angular velocity noise level ([m/s$^2$] and [\textdegree/s]) from $1.0 \times 10^{-3}$ to $1.0 \times 10^{-1}$. Although the environment was 40 m in width, the maximum observation range of the LiDAR was limited to 15 m so that range data became completely degenerated in the middle of the corridor for a few seconds. 

We ran the proposed odometry estimation algorithm and two state-of-the-art LiDAR-IMU odometry methods, namely LIO-SAM \cite{liosam2020shan}, a loosely coupled LiDAR-IMU odometry method, and FAST-LIO2 \cite{Xu2022}, a tightly coupled LiDAR-IMU odometry method with an iterated Kalman filter, to evaluate their robustness to range data degeneration. The estimated trajectories were evaluated with the absolute trajectory error (ATE \cite{Zhang2018}) metric using the {\it evo} toolkit\footnote{\url{https://michaelgrupp.github.io/evo/}}. Note that loop closure was disabled for all methods. 

Fig. \ref{fig:deg_trajs} shows the estimation results under the lowest IMU noise level ($1.0 \times 10^{-3}$). LIO-SAM and FAST-LIO, which are based on frame-to-model matching, are sensitive to the momentary degeneration of LiDAR data and thus their estimation results have large errors. Because frame-to-model matching needs to immediately determine and fix the sensor pose to merge the latest point cloud into the target map model, these methods cannot avoid corruption of the map model when point clouds are degenerated.

\begin{figure}[tb]
  \centering
  \subfloat[]{\includegraphics[width=0.8\linewidth]{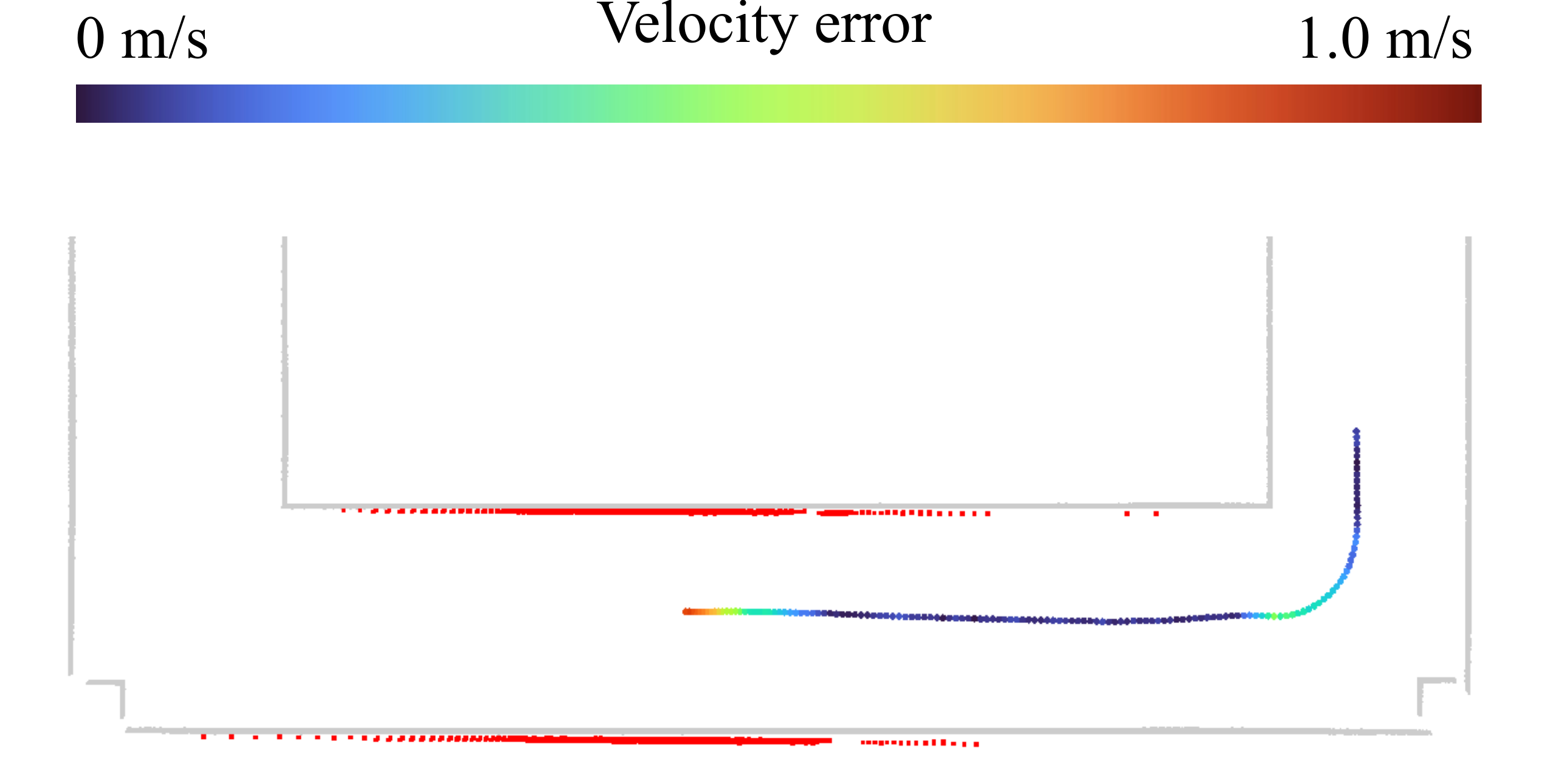}} \\
  \subfloat[]{\includegraphics[width=0.8\linewidth]{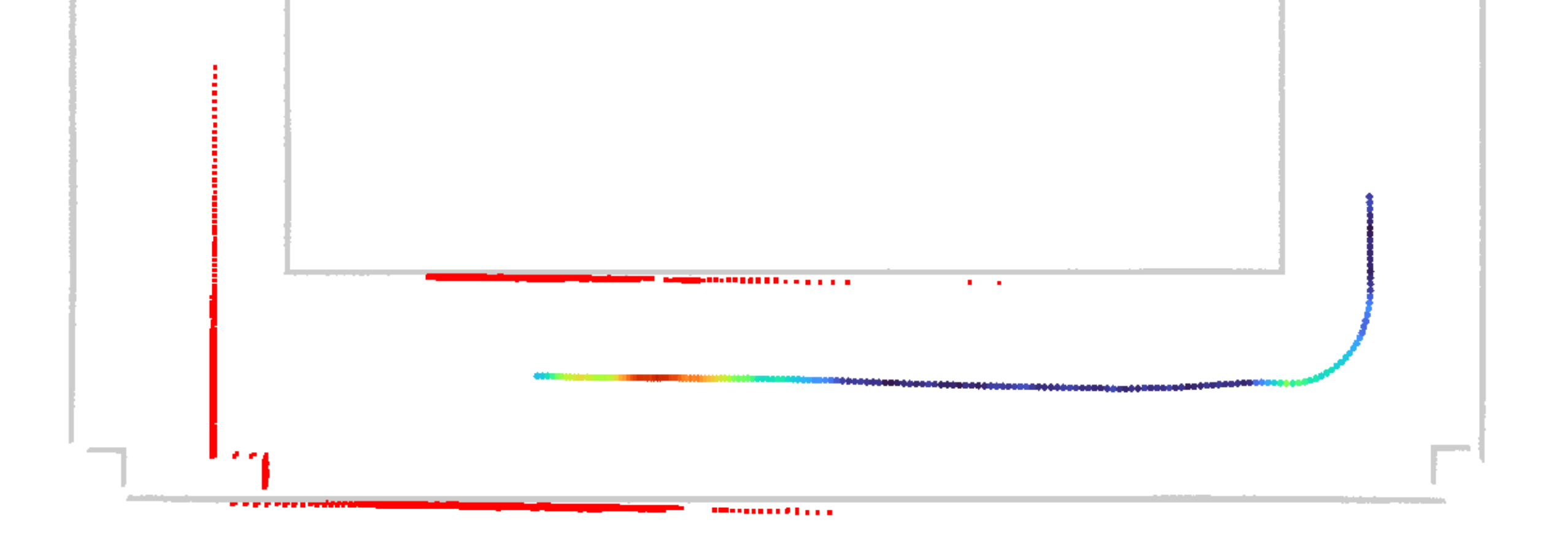}} \\
  \subfloat[]{\includegraphics[width=0.8\linewidth]{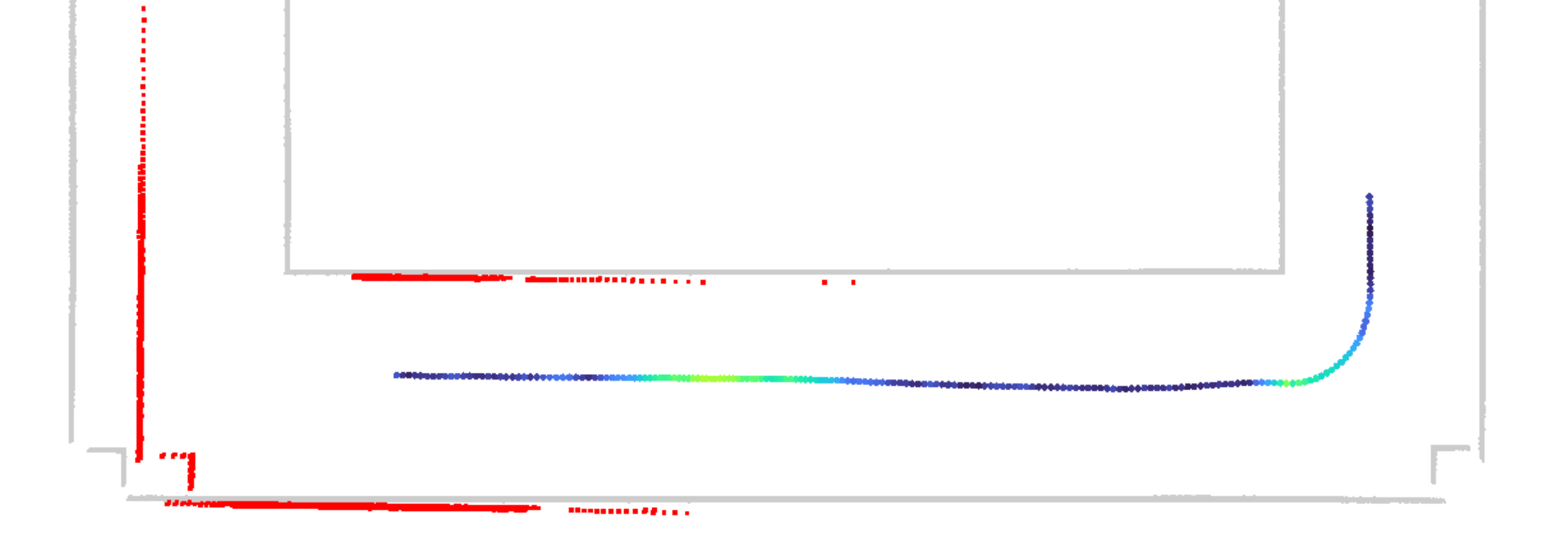}} \\
  \subfloat[]{\includegraphics[width=0.8\linewidth]{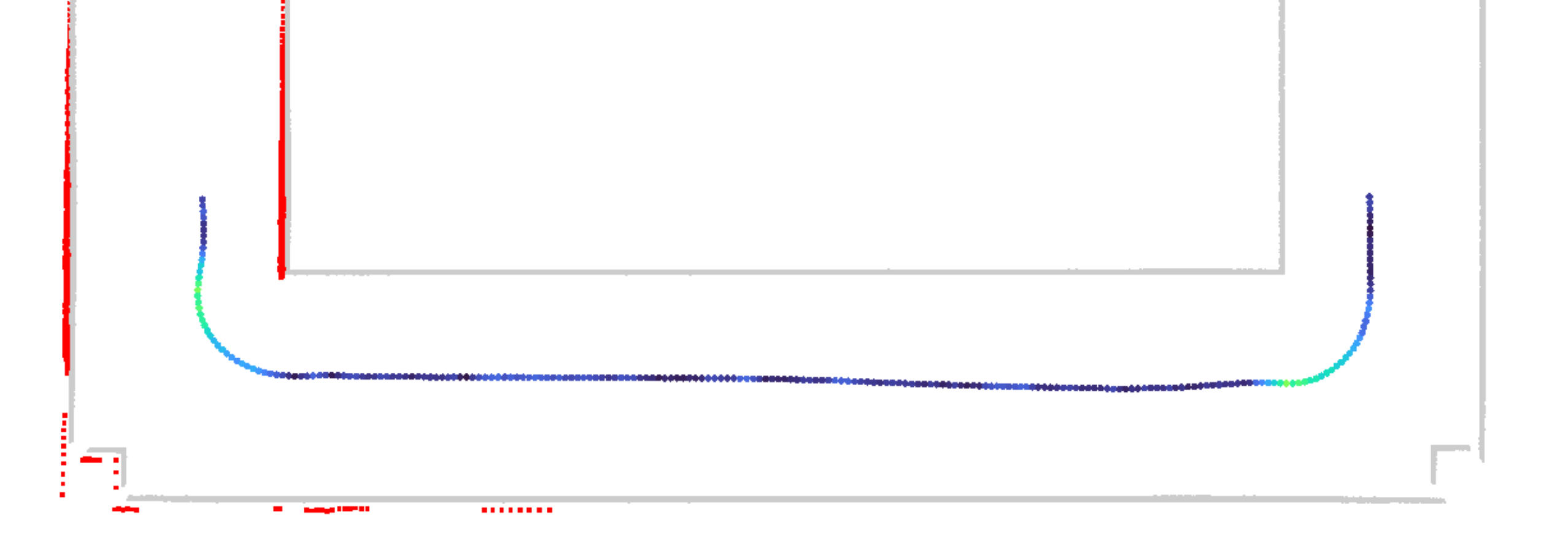}} \\
  \caption{Correction of velocity estimation error by proposed method for moving sensor. The trajectory color encodes the magnitude of the velocity estimation errors and the red points indicate the latest LiDAR scan. Map points shown in gray are shown to aid visualization. The estimation error accumulated in the middle of the corridor due to the degenerated LiDAR data. Once the opposite wall became observable, the estimation error of the latest frame was corrected. This correction was then propagated to past frames and a correct estimation was recovered at the end.}
  \label{fig:deg_vels}
\end{figure}

The proposed method had a small trajectory estimation error under degeneration of LiDAR data. Fig. \ref{fig:deg_vels} shows how the velocity estimation error for the proposed method was corrected while the sensor was moving in a corridor. The color of the trajectory encodes the magnitude of velocity errors and the red points show a 2D slice of the latest point cloud. While the velocity estimation error accumulated in the middle of the corridor due to the degenerated point clouds (Fig. \ref{fig:deg_vels} (a)), once the matching cost factors became well constrained by observing the opposite wall, the estimation of the latest frame was corrected (Fig. \ref{fig:deg_vels} (b)). It was then propagated to past frames (Fig. \ref{fig:deg_vels} (c)). The velocity estimates of all past frames were corrected and a correct trajectory estimation was recovered at the end (Fig. \ref{fig:deg_vels} (d)).


\begin{table}[tb]
  \centering
  \caption{Range Data Degeneration Test Results in Simulation}
  \label{tab:degenerated_sim}
  \begin{tabular}{g|ggg}
  \toprule
  \rowcolor{white}
  IMU Noise                               & \multicolumn{3}{c}{Absolute Trajectory Error [m]}  \\ \cmidrule{2-4}
  \rowcolor{white}
  $\text{[m/s}^2\text{] [\textdegree/s]}$ & LIO-SAM \cite{liosam2020shan}  & FAST-LIO2 \cite{Xu2022} & GLIM  \\
  \midrule
  \midrule
  \rowcolor{white}
  $1.0 \times 10^{-3}$ & 1.473 $\pm$ 0.696  & 1.294 $\pm$ 0.691  & \bf 0.099 $\pm$ 0.012 \\
  $5.0 \times 10^{-3}$ & 4.670 $\pm$ 1.863  & 1.449 $\pm$ 0.843  & \bf 0.064 $\pm$ 0.029 \\
  \rowcolor{white}
  $1.0 \times 10^{-2}$ & 12.888 $\pm$ 5.674 & 2.495 $\pm$ 1.111  & \bf 0.133 $\pm$ 0.083 \\
  $5.0 \times 10^{-2}$ & \xmark             & 11.735 $\pm$ 4.224 & \bf 0.384 $\pm$ 0.163 \\
  \rowcolor{white}
  $1.0 \times 10^{-1}$ & \xmark             & \xmark             & \bf 1.566 $\pm$ 0.839 \\
  \bottomrule
  \end{tabular}

  \vspace{2mm}
  \xmark \ indicates that the estimation was corrupted.
\end{table}

Table \ref{tab:degenerated_sim} summarizes the ATEs for each method under several IMU noise levels. Even under the lowest IMU noise level ($1.0 \times 10^{-3}$), LIO-SAM and FAST-LIO2 were sensitive to degenerated LiDAR data and showed large ATEs (1.474 and 1.294 m, respectively). As the IMU noise became larger, the ATEs for these methods also became larger. The output of LIO-SAM became corrupted when the noise level was set to $5.0 \times 10^{-2}$. Although FAST-LIO2, which employs iterated-Kalman-filter-based LiDAR-IMU tight coupling, was more robust to IMU noise compared to LIO-SAM, its output became corrupted under the highest IMU noise level ($1.0 \times 10^{-1}$).

The proposed method showed the best ATEs among the compared methods under all IMU noise level settings (0.099 to 1.566 m) thanks to its combination of fixed-lag smoothing and keyframe-based point cloud matching. Although its estimation error gradually increased as the IMU noise became larger, the proposed method had a reasonable ATE (1.566 m) even under the highest IMU noise level ($1.0 \times 10^{-1}$).

\begin{figure}[tb]
  \centering
  \subfloat[Environment]{\includegraphics[height=3.1cm]{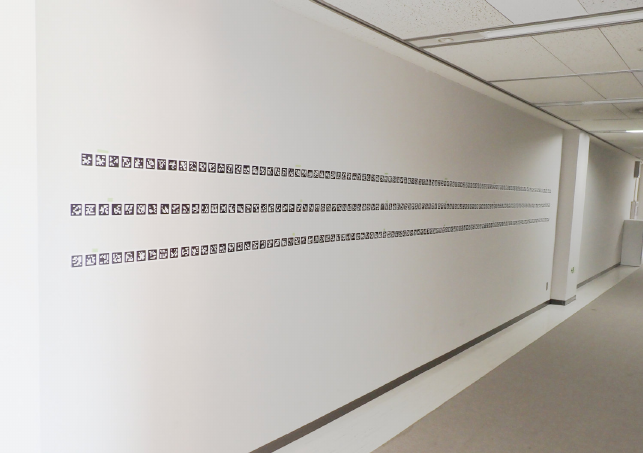}} \quad
  \subfloat[Used sensor]{\includegraphics[height=3.1cm]{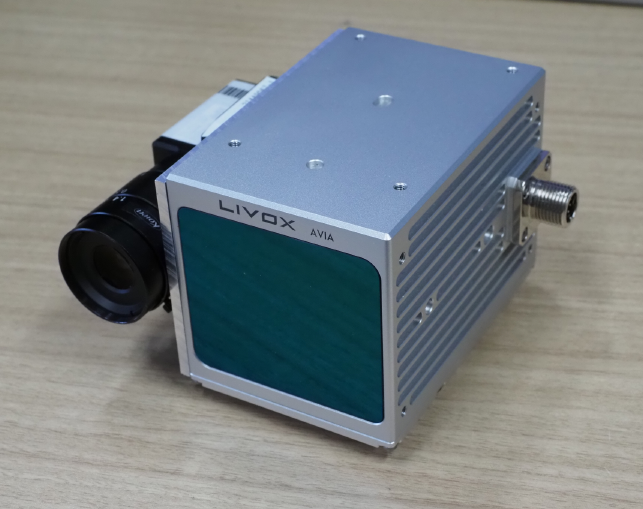}}
  \caption{Experimental environment and used sensors. The LiDAR system was moved between two pillars while facing the flat wall so the LiDAR data became degenerated in the middle of the path. The fiducial tags and camera data were used for acquiring ground truth trajectories.}
  \label{fig:flatwall}
\end{figure}


\begin{table*}[tb]
  \centering
  \caption{Range Data Degeneration Test Results in Real Environment}
  \label{tab:degenerated_real}
  \begin{tabular}{g|g|g|g|g|ggg}
  \toprule
  \rowcolor{white}
                            &          &  & \multicolumn{2}{c|}{Velocity (Avg / Max)} & \multicolumn{3}{c}{Absolute Trajectory Error [m]}  \\ \cmidrule{4-8}
  \rowcolor{white}
  \multirow{-2}[1]{*}{Seq.} & \multirow{-2}[1]{*}{Path length [m]} & \multirow{-2}[1]{*}{Duration [s]} & Linear [m/s] & Angular [\textdegree/s] & FAST-LIO2 \cite{Xu2022} & VoxelMap \cite{Yuan_2022} & GLIM        \\
  \midrule
  \midrule
  \rowcolor{white}
  01   & 4.106           & 18.5 & 0.22 / 0.55 & 11.6 / 39.9 & 0.815 $\pm$ 0.500  & 0.577 $\pm$ 0.157     & \bf 0.118 $\pm$ 0.047 \\
  02   & 4.192           & 13.6 & 0.32 / 0.62 & 13.4 / 33.6 & 0.822 $\pm$ 0.255  & \bf 0.146 $\pm$ 0.057 & 0.299 $\pm$ 0.116 \\
  \rowcolor{white}
  03   & 4.766           & 14.4 & 0.33 / 0.65 & 18.0 / 43.7 & 0.873 $\pm$ 0.364  & 0.950 $\pm$ 0.263     & \bf 0.040 $\pm$ 0.015 \\
  04   & 4.213           & 13.1 & 0.33 / 0.62 & 14.1 / 51.3 & 1.137 $\pm$ 0.412  & 0.586 $\pm$ 0.299     & \bf 0.389 $\pm$ 0.145 \\
  \rowcolor{white}
  05   & 4.121           & 14.0 & 0.30 / 0.69 & 17.1 / 50.0 &  1.048 $\pm$ 0.582  & 0.786 $\pm$ 0.400    & \bf 0.228 $\pm$ 0.083 \\
  06   & 3.725           & 8.8  & 0.43 / 0.79 & 16.0 / 42.7 &  15.551 $\pm$ 8.780 & 0.807 $\pm$ 0.309    & \bf 0.056 $\pm$ 0.023 \\
  \rowcolor{white}
  07   & 3.086           & 8.9  & 0.35 / 0.64 & 17.6 / 53.1 &  0.635 $\pm$ 0.218  & 0.366 $\pm$ 0.111    & \bf 0.017 $\pm$ 0.007 \\
  08   & 2.229           & 12.1 & 0.19 / 0.49 & 14.2 / 31.7 &  0.297 $\pm$ 0.110  & 0.279 $\pm$ 0.082    & \bf 0.146 $\pm$ 0.066 \\
  \bottomrule
  \end{tabular}
\end{table*}

{\bf Evaluation in real environment: } To show that the proposed method can deal with range data degeneration in real situations, we recorded eight LiDAR-IMU data sequences in the environment shown in Fig. \ref{fig:flatwall} with a non-repetitive scan LiDAR (Livox Avia). The LiDAR was moved between two pillars while facing the flat wall, and thus the LiDAR data were degenerated in the middle of the path for each sequence. To obtain the ground truth sensor trajectories, we placed fiducial tags (AprilTag \cite{Wang2016}) on the wall and estimated their poses using full bundle adjustment in advance. For each sequence, we recorded images using a camera (OMRON SENTECH STC-MBS202POE) along with LiDAR-IMU data. We estimated the camera trajectory via batch optimization of the tag projection and IMU constraints as a ground truth trajectory. The LiDAR and camera were synchronized with the host PC via the IEEE 1588 protocol (PTP) and the transformation between them was calibrated using a normalized-information-distance-based direct LiDAR-camera alignment method \cite{direct_visual_lidar}.

In addition to LIO-SAM \cite{liosam2020shan} and FAST-LIO2 \cite{Xu2022}, we ran VoxelMap \cite{Yuan_2022}, a tightly coupled LiDAR-IMU odometry estimation with adaptive resolutional voxel mapping.

Table \ref{tab:degenerated_real} summarizes the ATEs for the proposed method and FAST-LIO2 and VoxelMap. Note that we could not obtain decent estimation results with LIO-SAM \cite{liosam2020shan}. This was likely due to LIO-SAM employing a loosely coupled LiDAR-IMU fusion approach, in which point cloud scan matching is separately performed without IMU constraints. LIO-SAM thus has difficulty handling severely degenerated LiDAR data. From Table \ref{tab:degenerated_real}, we can see that FAST-LIO2 produced worse estimation results (0.297 to 15.551 m). Although VoxelMap showed better estimation results compared to those of FAST-LIO2 (0.146 to 0.950 m) thanks to its adaptive environment representation and data association, it also largely suffered from degenerated LiDAR data.

This reveals the weakness of state-of-the-art methods based on frame-to-model matching. As discussed in \cite{Xu2022}, FAST-LIO2 (and other methods based on frame-to-model matching) cannot deal with situations where point clouds are completely degenerated (e.g., when the LiDAR is facing a flat wall, the sky, or the ground or is close to objects).

The proposed method had better ATEs than those for FAST-LIO2 and VoxelMap for most of the sequences (0.017 to 0.389 m). We refer the reader to the supplementary page, which clearly demonstrates how the proposed fixed-lag smoothing approach behaves and corrects estimation drift by actively updating past sensor poses. 

\subsection{Mapping with Various Range-IMU Sensors}
\label{sec:exp_various}

\begin{figure}[tb]
  \centering
  \subfloat[Experimental environment]{\includegraphics[width=1.0\linewidth]{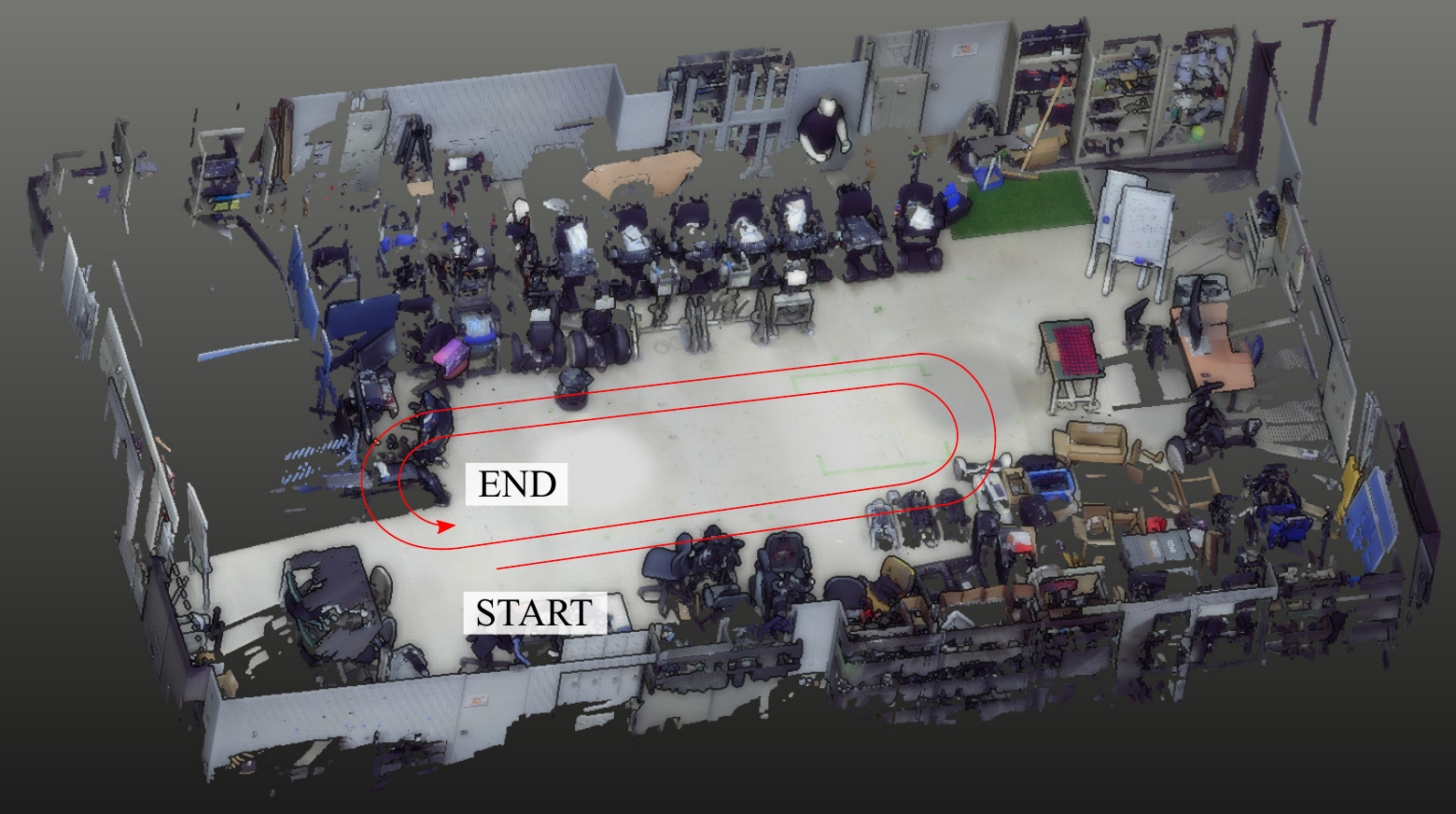}} \\
  \subfloat[Range-IMU sensors]{\includegraphics[width=1.0\linewidth]{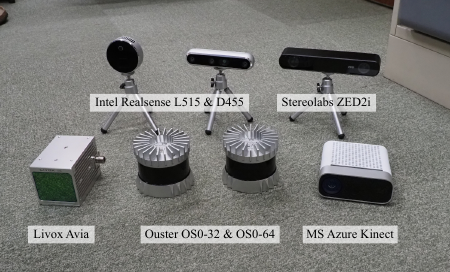}} \\
  \caption{Experimental environment and used range-IMU sensors. The proposed method was validated with various range-IMU sensors with different sensing mechanisms.}
  \label{fig:various}
\end{figure}


\begin{table}[tb]
  \centering
  \caption{Cross Sensor Evaluation}
  \label{tab:versatile}
  \begin{tabular}{g|gg}
  \toprule
  \rowcolor{white}
  Sensor           & ATE [m]              & RTE [m]  \\
  \midrule
  \rowcolor{white}
  Ouster OS0-32          & 0.037 $\pm$ 0.018 & 0.037 $\pm$ 0.017 \\
  Ouster OS0-64          & 0.022 $\pm$ 0.009 & 0.025 $\pm$ 0.012 \\
  \rowcolor{white}
  Livox Avia             & 0.041 $\pm$ 0.014 & 0.043 $\pm$ 0.019 \\
  Microsoft Azure Kinect & 0.007 $\pm$ 0.003 & 0.007 $\pm$ 0.003 \\
  \rowcolor{white}
  Intel Realsense L515   & 0.042 $\pm$ 0.018 & 0.045 $\pm$ 0.020 \\
  Intel Realsense D455   & 0.206 $\pm$ 0.086 & 0.331 $\pm$ 0.161 \\
  \rowcolor{white}
  Stereolabs ZED2i       & 0.139 $\pm$ 0.062 & 0.194 $\pm$ 0.102 \\
  \bottomrule
  \end{tabular}

  \vspace{2mm}
  ATE: absolute trajectory error; RTE: relative trajectory error
\end{table}

\begin{figure*}[tb]
  \centering
  \includegraphics[width=1.0\linewidth]{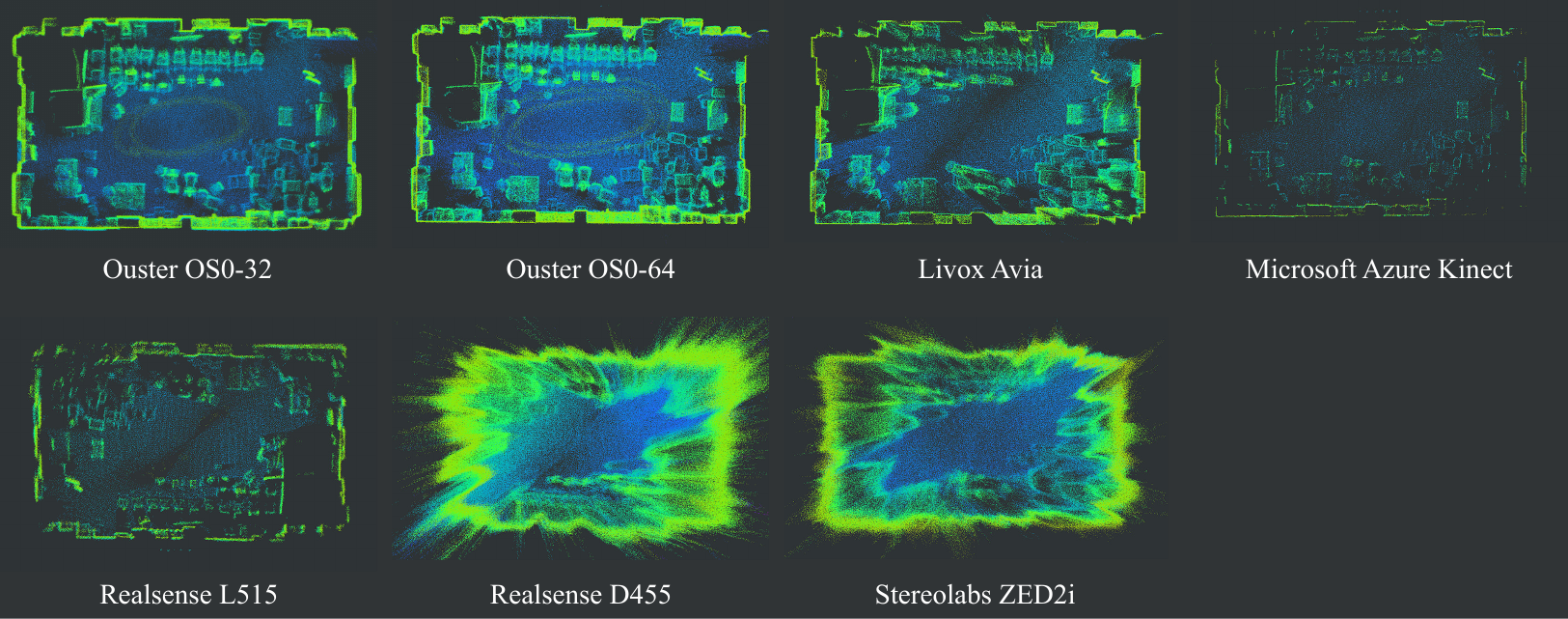}
  \caption{Mapping results obtained with various range-IMU sensors. The proposed method achieved consistent mapping results for all sensors.  The same mapping parameter set was used for all sensors.}
  \label{fig:various_maps}
\end{figure*}

To demonstrate the versatility of the proposed framework, we conducted mapping experiments with various range-IMU sensors with different sensing mechanisms including spinning LiDAR (Ouster OS0-32 and OS0-64), non-repetitive scan LiDAR (Livox Avia), time-of-flight (ToF) depth camera (Microsoft Azure Kinect), solid-state LiDAR (Intel Realsense L515), active stereo camera (Intel Realsense D455), and passive stereo camera (Stereolabs ZED2i). Fig. \ref{fig:various} shows the experimental environment and the used range-IMU sensors. Similar to the ground truth acquisition protocol used in the Newer College Dataset \cite{Ramezani2020}, we estimated sensor trajectories by aligning point clouds with an environment point cloud recorded with a survey-grade LiDAR (FARO Focus). 

We applied the proposed method to the recorded range-IMU sequences and evaluated the trajectory estimation errors using the ATE and relative trajectory error (RTE) metrics \cite{Zhang2018}. We set the sub-trajectory length of the RTE evaluation to 2 m. To show the generality of the proposed algorithm, we used the same mapping parameter set for all sensors. Note that we also tested FAST-LIO2 \cite{Xu2022}; however, we could not obtain reasonable results for sensors other than the Ouster and Livox LiDARs.

Table \ref{tab:versatile} summarizes the trajectory estimation errors. The proposed method showed accurate estimation results for the LiDARs and the ToF depth camera (OS0-32, OS0-64, Livox Avia, Azure Kinect, Realsense L515), which provided accurate point cloud data (ATE: 0.007 to 0.042 m, RTE: 0.007 to 0.045 m). Although the estimation results slightly deteriorated for the stereo-based cameras (Realsense D455 and Stereolabs ZED2i) due to strongly distorted point clouds, the estimation errors were still at a reasonable level (ATE: 0.139 and 0.206 m, RTE: 0.194 and 0.331 m).

Fig. \ref{fig:various_maps} shows the mapping results. For all sensors, the proposed method achieved consistent mapping results. Although the mapping results for stereo-based sensors were distorted because of the large distortion of the input point clouds, we obtained consistent maps without doubled walls and floors.

These results show the robustness of the proposed odometry estimation algorithm to noise on point clouds. Conventional odometry estimation methods based on feature-based point cloud matching are sensitive to such large noise in input point clouds, which makes the extraction of edge and plane feature points difficult. Because the proposed odometry estimation algorithm is based on direct distribution-to-distribution point cloud matching, which avoids noise-sensitive feature extraction and employs probabilistic surface modeling, it can robustly deal with noisy point clouds.

Furthermore, these results demonstrate the flexibility of the proposed global optimization algorithm based on global matching cost minimization. Because we cannot expect reasonable frame-by-frame scan matching results for the noisy point clouds of stereo-based sensors, existing methods based on pose graph optimization have difficulty accurately closing loops. In contrast, the global matching cost minimization approach does not explicitly require the relative pose between frames for each factor. Because it directly minimizes multi-scan registration errors on the factor graph, it is robust to such frame-by-frame matching failures.

\subsection{Quantitative Estimation Accuracy Evaluation on the Newer College Dataset}
\label{sec:exp_newer}

\begin{table*}[tb]
  \centering
  \caption{Data Durations and Path Lengths for Multi-Camera Newer College Dataset}
  \label{tab:newer_duration}
  \begin{tabular}{g|ggggggggg}
  \toprule
  \rowcolor{white}                             & quad  & quad   & quad  &                         &                         &                            & math  & math   & math \\
  \rowcolor{white} \multirow{-2}{*}{Sequence} & easy  & medium & hard  &\multirow{-2}{*}{stairs} &  \multirow{-2}{*}{park} & \multirow{-2}{*}{cloister} & easy  & medium & hard \\
  \midrule
  \midrule
  \rowcolor{white} Duration [s]          & 198.7 & 190.6  & 187.8 & 118.9 & 1572.0 & 278.6 & 215.9 & 176.9  & 243.7 \\
                   Path length [m]       & 246.7 & 260.4  & 234.8 & 57.0  & 2396.2 & 428.8 & 263.6 & 304.3  & 320.6 \\
  \bottomrule
  \end{tabular}
\end{table*}

\begin{table*}[tb]
  \centering
  \caption{Absolute Trajectory Error [m] Results for Various Methods on Multi-Camera Newer College Dataset}
  \label{tab:newer_ape}
  \begin{tabular}{g|g|ggggggggg|g}
  \toprule
  \rowcolor{white}                           & Loop     & quad & quad   & quad &                          &                        &                            & math & math   & math &                           \\
  \rowcolor{white} \multirow{-2}{*}{Method}  & closure  & easy & medium & hard & \multirow{-2}{*}{stairs} & \multirow{-2}{*}{park} & \multirow{-2}{*}{cloister} & easy & medium & hard & \multirow{-2}{*}{Average} \\
  \midrule
  \midrule
  \rowcolor{white} LINS \cite{Qin2020}          &            & 0.160      & 0.212      & 16.824      & 3.405      & 0.612      & 1.170      & 0.216      & 0.259      & 5.710      & 3.174 \\
                                                &            & 0.086      & \bf 0.069  & 0.105       & 3.438      & 1.381      & 0.085      & \bf 0.088  & \bfb 0.114 & 0.089      & 0.606 \\
  \multirow{-2}{*}{LIO-SAM\cite{liosam2020shan}}& \checkmark & 0.083      & 0.075      & 0.099       & 2.942      & \bf 0.519  & 0.082      & \bfb 0.083 & \bfr 0.111 & 0.092      & 0.454 \\
  \rowcolor{white} FAST-LIO2 \cite{Xu2022}      &            & \bfr 0.068 & \bfr 0.059 & \bf 0.050   & 1.320      & \bfb 0.319 & \bf 0.078  & 0.091      & 0.134      & \bfb 0.058 & 0.242 \\
                                                &            & 0.197      & 0.674      & 0.151       & 2.336      & 18.265     & 0.293      & 0.225      & 0.235      & 0.494      & 2.541 \\
  \multirow{-2}{*}{CLINS \cite{clins}}          & \checkmark & 0.252      & 0.451      & 0.140       & 2.321      & 2.757      & 0.262      & 0.241      & 0.242      & 0.484      & 0.795 \\
  \rowcolor{white} DLO \cite{Chen2022}          &            & \bf 0.080  & 0.100      & 0.168       & \bf 0.129  & 0.686      & 0.163      & 0.161      & 0.211      & 0.192      & \bf 0.210 \\
  \midrule
  
                                                &            & \bfb 0.070 & \bfb 0.061 & \bfb 0.044  & \bfb 0.106 & 0.459      & \bfb 0.063 & 0.096      & \bf 0.119  & \bf 0.064  & \bfb 0.120 \\
  \multirow{-2}{*}{GLIM}                        & \checkmark & \bfb 0.070 & \bfr 0.059 & \bfr 0.043  & \bfr 0.046 & \bfr 0.269 & \bfr 0.056 & \bfr 0.082 & \bf 0.119  & \bfr 0.048 & \bfr 0.088 \\
  \bottomrule
  \end{tabular}

  \vspace{2mm}
  {\bfr Red}, {\bfb blue}, and {\bf black} bold values are respectively the {\bfr best}, {\bfb second best}, and {\bf third best} results. \\

  \vspace{5mm}
  \centering
  \caption{Relative Trajectory Error [m] Results for Various Methods on Multi-Camera Newer College Dataset}
  \label{tab:newer_rpe}
  \begin{tabular}{g|g|ggggggggg|g}
  \toprule
  \rowcolor{white}                           & Loop     & quad & quad   & quad &                          &                        &                            & math & math   & math &                           \\
  \rowcolor{white} \multirow{-2}{*}{Method}  & closure  & easy & medium & hard & \multirow{-2}{*}{stairs} & \multirow{-2}{*}{park} & \multirow{-2}{*}{cloister} & easy & medium & hard & \multirow{-2}{*}{Average} \\
  \midrule
  \midrule
  \rowcolor{white} LINS \cite{Qin2020}           &            & 0.328      & 0.463      & 0.996      & 4.592       & 0.435      & 0.923      & 0.306      & 0.523      & 0.887      & 1.050 \\
                                                 &            & \bfr 0.143 & 0.172      & 0.201      & 2.662       & 0.174      & 0.109      & 0.129      & \bf 0.223  & 0.246      & 0.451 \\
  \multirow{-2}{*}{LIO-SAM\cite{liosam2020shan}} & \checkmark & \bf 0.151  & 0.195      & 0.209      & 2.630       & 0.205      & 0.117      & \bfb 0.118 & \bfb 0.216 & 0.211      & 0.450 \\
  \rowcolor{white} FAST-LIO2 \cite{Xu2022}       &            & \bfb 0.150 & \bfr 0.161 & \bfr 0.152 & 1.137       & \bfr 0.090 & \bfb 0.103 & 0.146      & 0.293      & \bfr 0.102 & 0.259 \\
                                                 &            & 0.177      & 0.271      & 0.196      & 3.070       & 0.291      & 0.149      & 0.145      & 0.276      & \bfb 0.120 & 0.522 \\
  \multirow{-2}{*}{CLINS \cite{clins}}           & \checkmark & 0.202      & 0.486      & \bf 0.194  & 2.933       & 0.157      & 0.155      & 0.145      & 0.282      & 0.163      & 0.524 \\
  \rowcolor{white} DLO \cite{Chen2022}           &            & 0.222      & 0.328      & 0.365      & \bf 0.192   & 0.233      & 0.240      & \bfr 0.053 & \bfr 0.108 & 0.379      & \bf 0.236 \\
  \midrule
                                                 &            & \bf 0.151  & \bfb 0.168 & \bfb 0.164 & \bfb 0.169  & \bf 0.096  & \bfr 0.100 & 0.125      & 0.250      & 0.137      & \bfb 0.151 \\
  \multirow{-2}{*}{GLIM}                         & \checkmark & 0.154      & \bf 0.170  & \bfb 0.164 & \bfr 0.096  & \bfb 0.095 & \bf 0.107  & \bf 0.123  & 0.243      & \bf 0.135  & \bfr 0.143 \\
  \bottomrule
  \end{tabular}

  \vspace{2mm}
  {\bfr Red}, {\bfb blue}, and {\bf black} bold values are respectively the {\bfr best}, {\bfb second best}, and {\bf third best} results. \\
\end{table*}

\begin{figure*}[tb]
 \centering
 \includegraphics[width=0.99\linewidth]{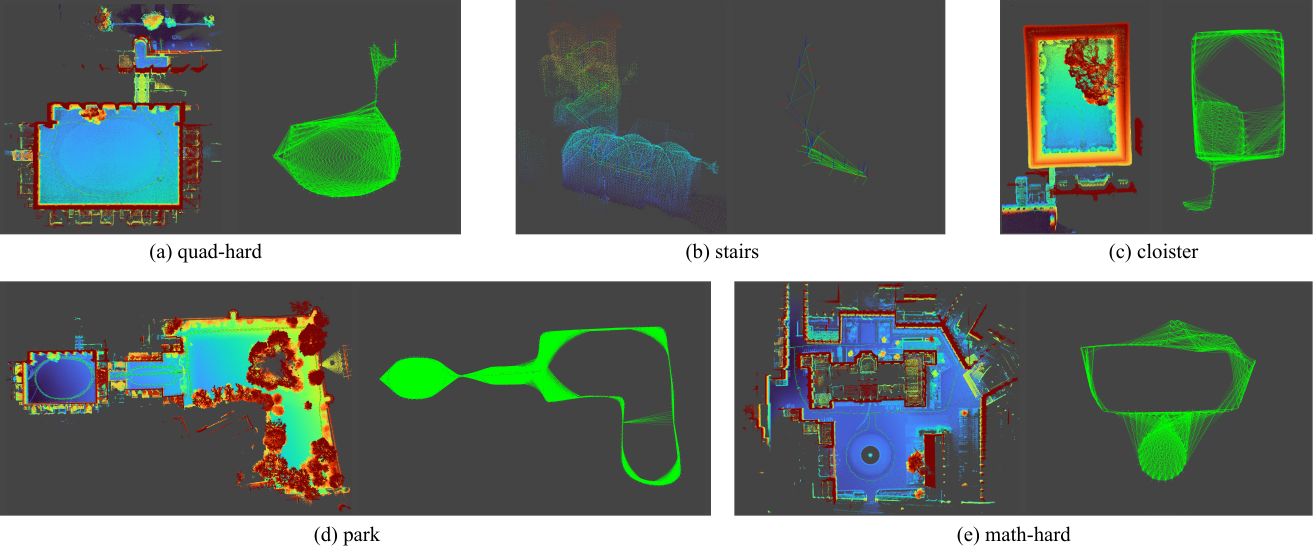}
 \caption{Mapping results and factor graphs for Multi-Camera Newer College Dataset. Highly consistent environmental maps were created via global matching cost minimization on extremely dense factor graphs.}
 \label{fig:newer_results}
\end{figure*}


\begin{table}[tb]
  \centering
  \caption{Processing Time for {\it park} Sequence}
  \label{tab:proctime}
  \begin{tabular}{llc}
  \toprule
  Module & Process & Time [ms] \\
  \midrule
  \midrule
  \multirow{3}{*}{Preprocess}
                      & Downsampling       & 8.9 $\pm$ 1.5  \\
                      & kNN search         & 5.7 $\pm$ 0.4 \\
                      & Total (per frame)  & 15.1 $\pm$ 1.6 \\
  \midrule
  \multirow{3}{*}{Odometry estimation}
                      & Factor creation    & 3.3 $\pm$ 0.3   \\
                      & Optimization       & 25.7 $\pm$ 30.5 \\
                      & Total (per frame)  & 29.3 $\pm$ 30.5 \\
  \midrule
  \multirow{4}{*}{Local mapping}
                      & Factor creation    & 12.3 $\pm$ 2.0    \\
                      & Total (per frame)  & 12.4 $\pm$ 2.0    \\ \cmidrule{2-3}
                      & Optimization       & 295.0 $\pm$ 111.0 \\
                      & Total (per submap) & 295.0 $\pm$ 111.0 \\
  \midrule
  \multirow{3}{*}{Global mapping}
                      & Factor creation    & 20.1 $\pm$ 10.3   \\
                      & Optimization       & 35.6 $\pm$ 51.4 \\
                      & Total (per submap) & 55.8 $\pm$ 55.6 \\
  \bottomrule
  \end{tabular}
\end{table}

\begin{figure}[tb]
 \centering
 \includegraphics[width=\linewidth]{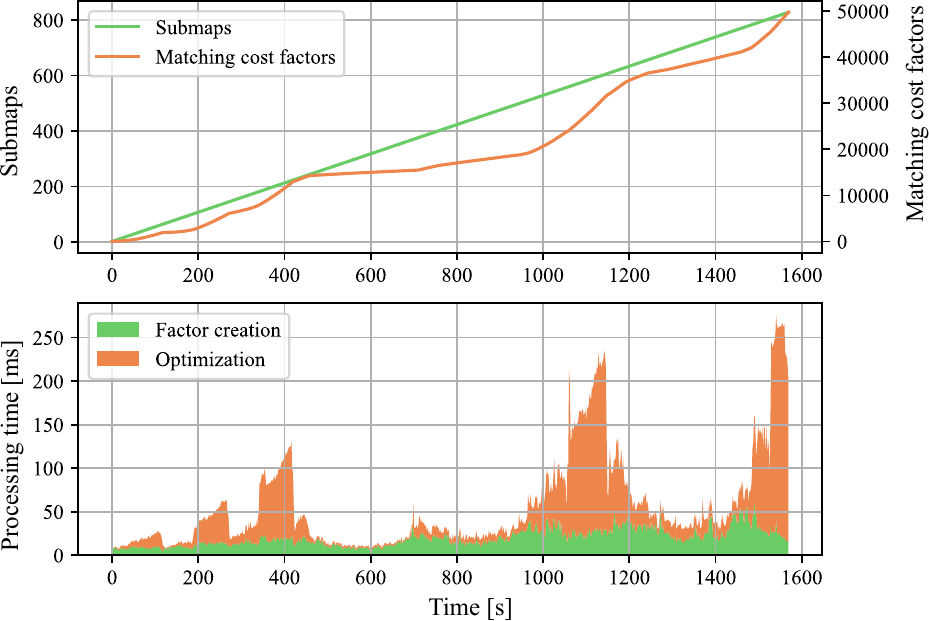}
 \caption{Number of submaps and matching cost factors and processing time of global optimization module for {\it park} sequence. Owing to the efficient and incremental optimization scheme, the proposed method achieved a real-time processing.}
 \label{fig:proctime}
\end{figure}

We quantitatively evaluated the estimation accuracy of the proposed framework on the Multi-Camera Newer College Dataset \cite{Ramezani2020,zhang2021multicamera}. This dataset provides sequences of LiDAR-camera-IMU data recorded with an Ouster OS0-128 (10-Hz point clouds and 100-Hz IMU) and a Sevensense Alphasense Core (30-Hz images $\times$ 4 hardware-synchronized cameras) in several indoor and outdoor environments. The LiDAR and cameras were synchronized via the IEEE 1588 protocol (PTP). Table \ref{tab:newer_duration} summarizes the data duration and path length of each sequence in the dataset.

{\bf Comparison with state-of-the-art:}
We compared the trajectory estimation accuracy of the proposed method with those of state-of-the-art LiDAR-IMU SLAM methods, including a loosely coupled LiDAR-IMU method (LIO-SAM \cite{liosam2020shan}), tightly coupled methods based on an iterated Kalman filter (LINS \cite{Qin2020}, FAST-LIO2 \cite{Xu2022}), a tightly coupled continuous SLAM method (CLINS \cite{clins}), and a loosely coupled method based on GICP matching (DLO \cite{Chen2022}). Because LINS, FAST-LIO2, and DLO do not have a loop closure mechanism, only odometry estimation errors were evaluated for these methods. We evaluated the estimation errors with the translational ATE and RTE (delta = 10 m) metrics.

Tables \ref{tab:newer_ape} and \ref{tab:newer_rpe} respectively summarize the ATEs and RTEs for the evaluated methods. Among the existing methods, FAST-LIO2 showed the best ATEs and RTEs for most of the sequences owing to the robust state estimation based on tightly coupled range and IMU data fusion on an iterated Kalman filter. However, the ATEs and RTEs for FAST-LIO2 and the other methods based on point-to-plane (and point-to-edge) point cloud matching (LINS, LIO-SAM, and CLINS) deteriorated significantly for the {\it stairs} sequence in a small indoor environment. 

We consider this to be due to point-to-plane-based matching requiring precise point association and needing careful tuning of parameters to deal with environment changes. It is worth mentioning that although we fine-tuned the parameters of these methods to improve the results on the {\it stairs} sequence, we could not obtain an improvement without sacrificing accuracy for the other sequences in outdoor environments. DLO showed better ATE and RTE for the {\it stairs} sequence because it employs a direct scan matching algorithm based on GICP, which does not require feature extraction and is robust to environment changes. However, the overall accuracy of DLO was slightly worse than that of FAST-LIO2.

The proposed odometry estimation algorithm showed comparable ATEs and RTEs to those for FAST-LIO2 for most of the sequences. For the {\it stairs} sequence, it showed the best estimation result among all evaluated methods. This suggests that the proposed method can deal with a large variety of environments while suppressing estimation drift. With loop closure, the ATEs and RTEs for the proposed method were further improved, and we achieved the best ATEs for most of the sequences. Fig. \ref{fig:newer_results} shows the map point clouds and factor graphs created with the proposed framework. We can see that the proposed method created an extremely dense factor graph for global matching cost minimization, which resulted in consistent mapping results.

\begin{table*}[tb]
  \centering
  \caption{Absolute Trajectory Error [m] Results for Odometry Estimation Ablation Study}
  \label{tab:newer_abl_frontend}
  \begin{tabular}{g|ggggggggg|g}
  \toprule
  \rowcolor{white}                           & quad & quad   & quad &                          &                        &                            & math & math   & math &                           \\
  \rowcolor{white} \multirow{-2}{*}{Method}  & easy & medium & hard & \multirow{-2}{*}{stairs} & \multirow{-2}{*}{park} & \multirow{-2}{*}{cloister} & easy & medium & hard & \multirow{-2}{*}{Average} \\
  \midrule
  \midrule
  \rowcolor{white} GLIM odometry (baseline)                    & 0.070         & 0.061         & 0.044         & 0.106         & 0.459         & 0.063         & 0.096         & 0.119         & 0.064    & 0.120 \\
                                                               & 0.070         & 0.061         & \bfr 0.043    & \bfb 0.114    & \bfb 0.533    & \bfr 0.061    & \bfb 0.140    & \bfb 0.147    & \bfb 0.097    & \bfb 0.140 \\
  \multirow{-2}{*}{w/o Surface validation}                     & (+0.000)      & (+0.000)      & \bfr (-0.001) & \bfb (+0.008) & \bfb (+0.074) & \bfr (-0.002) & \bfb (+0.044) & \bfb (+0.027) & \bfb (+0.033) & \bfb (+0.020) \\
  \rowcolor{white}                                             & \bfb 0.078    & \bfb 0.092    & \bfb 0.071    & \bfr 0.073    & \bfb 1.096    & \bfb 0.355    & \bfb 0.127    & \bfb 0.140    & \bfb 0.074    & \bfb 0.234    \\
  \rowcolor{white} \multirow{-2}{*}{w/o Multi resolution}      & \bfb (+0.008) & \bfb (+0.031) & \bfb (+0.027) & \bfr (-0.033) & \bfb (+0.637) & \bfb (+0.292) & \bfb (+0.031) & \bfb (+0.021) & \bfb (+0.010) & \bfb (+0.114) \\
                                                               & \bfr 0.069    & \bfr 0.059    & \bfr 0.040    & \bfr 0.045    & \bfr 0.418    & \bfr 0.053    & \bfr 0.095    & \bfb 0.122    & \bfr 0.055    & \bfr 0.106    \\
  \multirow{-2}{*}{w/ Multi-cam constraints}                   & \bfr (-0.001) & \bfr (-0.002) & \bfr (-0.004) & \bfr (-0.061) & \bfr (-0.041) & \bfr (-0.010) & \bfr (-0.001) & \bfb (+0.003) & \bfr (-0.009) & \bfr (-0.014) \\
  \bottomrule
  \end{tabular}

  \vspace{2mm}
  {\bfr Red} and {\bfb blue} values are respectively {\bfr improved} and {\bfb deteriorated} results. \\
  Values in parentheses indicate absolute trajectory error differences from those of the baseline.
\end{table*}

\begin{table*}[tb]
  \centering
  \caption{Absolute Trajectory Error [m] Results for Global Optimization Ablation Study}
  \label{tab:newer_abl_backend}
  \begin{tabular}{g|ggggggggg|g}
  \toprule
  \rowcolor{white}                           & quad & quad   & quad &                          &                        &                            & math & math   & math &                           \\
  \rowcolor{white} \multirow{-2}{*}{Method}  & easy & medium & hard & \multirow{-2}{*}{stairs} & \multirow{-2}{*}{park} & \multirow{-2}{*}{cloister} & easy & medium & hard & \multirow{-2}{*}{Average} \\
  \midrule
  \midrule
  \rowcolor{white} GLIM         (Baseline)                  & 0.070    & 0.059    & 0.043         & 0.046         & 0.269         & 0.056         & 0.082         & 0.119    & 0.048    & 0.088    \\
                                                            & 0.070    & 0.059    & \bfb 0.045    & \bfb 0.054    & \bfr 0.268    & \bfb 0.217    & \bfr 0.081    & 0.119    & 0.048    & \bfb 0.107    \\
                   \multirow{-2}{*}{w/o IMU factors}    & (+0.000) & (+0.000) & \bfb (+0.002) & \bfb (+0.008) & \bfr (-0.001) & \bfb (+0.161) & \bfr (-0.001) & (+0.000) & (+0.000) & \bfb (+0.019) \\
  \rowcolor{white}                                          & 0.070    & 0.059    & 0.043         & \bfb 0.054    & \bfr 0.268    & \bfb 0.341    & \bfb 0.082    & 0.119    & 0.048    & \bfb 0.120    \\
  \rowcolor{white} \multirow{-2}{*}{w/ IMU factors between submaps} & (+0.000) & (+0.000) & (+0.000)      & \bfb (+0.008) & \bfr (-0.001) & \bfb (+0.285) & \bfb (+0.000) & (+0.000) & (+0.000) & \bfb (+0.032) \\
  \bottomrule
  \end{tabular}

  \vspace{2mm}
  {\bfr Red} and {\bfb blue} values are respectively {\bfr improved} and {\bfb deteriorated} results. \\
  Values in parentheses indicate absolute trajectory error differences from those of the baseline.
\end{table*}

{\bf Processing time:} 
Table \ref{tab:proctime} summarizes the processing times, measured with an Intel Core i7 8700K and an NVIDIA RTX 1660 Ti, for each module in the proposed framework running through the {\it park} sequence, which is the longest sequence in the dataset.

The preprocessing and odometry estimation modules respectively took 15.1 and 29.3 ms per frame on average and are thus sufficiently faster than the real-time requirement (100 ms per frame). The submap optimization, which was performed approximately every 2 s, took 295.0 ms on average. The global map optimization took 55.8 ms on average for each submap creation. Fig. \ref{fig:proctime} shows how the global optimization time grew as the number of submaps and matching cost factors increased. Although a massive number of matching cost factors were created (over 50,000 factors between 800 submaps), the global optimization converged in less than 50 ms for most of the frames thanks to the GPU-accelerated matching cost factor and iSAM2, which efficiently updates only sub Bayes trees that can be influenced by the insertion of a new submap. Although the global optimization took a longer time when closing a large loop, the maximum optimization time was only about 250 ms, which is much smaller than the real-time requirement (2 s). Note that the linearization of the matching cost factors accounted for most of the optimization time; the linear solver performed on the CPU accounted for only about 5\% of the total optimization time.

{\bf Odometry estimation ablation study:}
To examine the effects of surface-orientation-based correspondence validation and a multi-resolution voxelmap, we ran the proposed framework using the same settings used in the previous experiment but without these functionalities. We also ran the proposed odometry estimation algorithm with the multi-camera constraints to show that we can improve the mapping accuracy in real situations by introducing additional constraints via the global callback slot mechanism. Table \ref{tab:newer_abl_frontend} summarizes the evaluation results.

Without the surface-orientation-based correspondence validation, the ATEs deteriorated for about half of the sequences ({\it stairs, park, math-easy, math-medium}, and {\it math-hard}) and the average ATE increased from 0.120 to 0.140 m. We consider this to be due to the points on a surface being possibly wrongly associated with points on the other side of the surface in another frame when correspondence validation is not performed.

Without the multi-resolution voxelmap, we observed an accuracy improvement for the {\it stairs} sequence (ATE decreased from 0.106 to 0.073 m). However, for all other sequences, the ATEs largely deteriorated, resulting in a deteriorated average ATE (0.234 m). The ATEs tend to be large for large outdoor environments (e.g., {\it park} and {\it cloister}). The results indicate that the multi-resolution voxelmap enables a robust association of points and voxels in a wide variety of environments at the cost of a slight accuracy decrease in small indoor environments.

With the multi-camera visual constraints, the ATEs improved for all sequences except for {\it math-medium} and the average ATEs decreased from 0.120 to 0.106 m. The ATE improvements were marginal in sequences where rich geometric features were available and scan matching was well constrained. The ATEs for the {\it stairs}, {\it park}, {\it cloister}, and {\it math-hard} sequences, where point cloud matching might be unstable due to environment scale changes and aggressive sensor motion, were largely improved with the tight coupling of visual constraints.

{\bf Global trajectory optimization ablation study:}
We evaluated the effect of the proposed {\it endpoint} mechanism for stabilizing submap pose optimization with IMU constraints. We ran the proposed framework with two settings: 1) without IMU constraints and 2) with IMU constraints created directly between submap states. Table \ref{tab:newer_abl_backend} summarizes the evaluation results. We can see that for both settings, the ATEs deteriorated significantly for the {\it cloister} sequence. When the framework was running on the {\it cloister} sequence, the optimizer had indeterminant linear system errors, which indicate that the factor graph was under-constrained. Even when we inserted IMU factors directly between submap states, the optimizer still suffered from an under-constrained system. This is because submaps were created with a large time interval and the IMU factors created between them had a large integration time, resulting in large uncertainty. This shows the advantage of the proposed {\it endpoint} mechanism, which can strongly constrain the submap poses because each IMU factor covers only a minimum scan interval by being created between the {\it endpoints} of consecutive submaps.

\subsection{Quantitative Estimation Accuracy Evaluation on the NTU VIRAL Dataset}
\label{sec:exp_viral}

We further compared the accuracy of the proposed method with those of state-of-the-art methods in a more challenging environment on the NTU VIRAL dataset \cite{Nguyen_2021}. This dataset provides time-synchronized sequences of two LiDARs (Ouster OS1-16), two cameras (uEye 1221 LE), an IMU (VectorNav VN100), and a UWB (Humatic P440) on a UAV. Because it was recorded with dynamic 3D motion of a flying UAV, we consider evaluation on this dataset would reveal the robustness of state-of-the-art methods in challenging situations.

We ran the proposed method with three configurations (LiDAR-IMU, visual-LiDAR-IMU, LiDAR-IMU with loop closure) and compared its accuracy with those of state-of-the-art LiDAR-IMU odometry estimation methods (LIO-SAM \cite{liosam2020shan}, MLOAM \cite{Jiao_2022}, FAST-LIO2 \cite{Xu2022}, VoxelMap \cite{Yuan_2022}, and BALM \cite{liu2020balm}), visual-LiDAR-IMU odometry estimation methods (VIRAL-SLAM \cite{viralslam} and FAST-LIVO \cite{Zheng_2022}), and a LiDAR-IMU mapping method with loop closure (SLICT \cite{slict}). We evaluated the translational ATEs of those methods by using the evaluation code provided by the dataset\footnote{\url{https://github.com/ntu-aris/viral_eval}}.

Table \ref{tab:viral} summarizes the ATEs for the evaluated methods. The ATEs for LIO-SAM, MLOAM, VIRAL-SLAM, and SLICT were taken from \cite{viralslam} and \cite{slict}.

Among the existing LiDAR-IMU-based methods, FAST-LIO2 showed accurate estimation results (Average ATE: 0.040 m) owing to its efficient fusion of LiDAR and IMU constraints. We can also see that BALM also showed comparable results (0.039 m) owing to its BA-based optimization ensuring consistent mapping results. The proposed method showed the best ATEs for six out of the nine sequences that resulted in the best average ATE (0.031 m) among LiDAR-IMU-based methods. We consider the fixed-lag-smoothing-based optimization enabled dealing with the UAV's dynamic motion robustly and achieved smooth and accurate trajectory estimation results.

With visual constraints, the proposed method improved the average ATE (0.030 m) that was the best results among the evaluated visual-LiDAR-IMU-based methods. This result confirms that the proposed tight fusion of visual constraints enable enhancing the estimation accuracy without sacrificing the accuracy of the LiDAR-IMU-based estimation.

Finally, with loop closure, the ATE of the proposed method got further improved to 0.025 m that was better than the ATE of SLICT (0.028 m). We observed large accuracy gain in large outdoor sequences that demonstrate the effectiveness of the proposed global trajectory optimization to compensate for odometry estimation drift in challenging situations.

\begin{table*}[tb]
  \centering
  \caption{Relative Trajectory Error [m] Results for Various Methods on NTU VIRAL Dataset}
  \label{tab:viral}
  \begin{tabular}{g|g|g|ggggggggg|g}
  \toprule
  \rowcolor{white}                                  &                          & Loop     & eee & eee & eee & nya & nya & nya & sbs & sbs & sbs &                           \\
  \rowcolor{white} \multirow{-2}{*}{Method}         & \multirow{-2}{*}{Camera} & closure  & 01  & 02  & 03  & 01  & 02  & 03  & 01  & 02  & 03  & \multirow{-2}{*}{Average} \\
  \midrule
  \midrule
  \multicolumn{13}{c}{LiDAR-based methods} \\
  \midrule
  \rowcolor{white} LIO-SAM \cite{liosam2020shan}    &            &            & 0.075     & 0.069     & 0.101     & 0.076     & 0.090     & 0.137     & 0.089     & 0.083     & 0.140     & 0.096 \\
                   MLOAM \cite{Jiao_2022} \dag      &            &            & 0.249     & 0.166     & 0.232     & 0.123     & 0.191     & 0.226     & 0.173     & 0.147     & 0.153     & 0.184 \\
  \rowcolor{white} FAST-LIO2 \cite{Xu2021}           &            &            & 0.056     & 0.031     & 0.047     & 0.041     & 0.047     & 0.036     & \bf 0.036 & 0.033     & \bf 0.034 & 0.040 \\
                   VoxelMap \cite{Yuan_2022}        &            &            & 0.075     & 0.033     & 0.070     & 0.048     & 0.040     & 0.050     & 0.041     & 0.072     & 0.058     & 0.054 \\
  \rowcolor{white} BALM \cite{liu2020balm}          &            &            & 0.060     & \bf 0.021 & \bf 0.028 & \bf 0.031 & 0.037     & 0.028     & 0.041     & 0.054     & 0.055     & 0.039 \\
                   GLIM                             &            &            & \bf 0.034 & 0.030     & 0.032     & \bf 0.031 & \bf 0.028 & \bf 0.027 & 0.037     & \bf 0.025 & \bf 0.034 & \bf 0.031 \\
  \midrule
  \midrule
  \multicolumn{13}{c}{Visual-LiDAR-based methods} \\
  \midrule
                   VIRAL-SLAM \cite{viralslam} \dag & \checkmark &            & 0.060     & 0.058     & 0.037     & 0.051     & 0.043     & 0.032     & 0.048     & 0.062     & 0.054     & 0.049 \\
  \rowcolor{white} FAST-LIVO \cite{Zheng_2022}      & \checkmark &            & 0.077     & \bf 0.020 & 0.032     & 0.034     & 0.044     & \bf 0.025 & 0.039     & 0.044     & 0.039     & 0.039 \\
                   GLIM                             & \checkmark &            & \bf 0.032 & 0.026     & \bf 0.030 & \bf 0.031 & \bf 0.028 & 0.026     & \bf 0.037 & \bf 0.025 & \bf 0.033 & \bf 0.030 \\
  \midrule
  \midrule
  \multicolumn{13}{c}{LiDAR-based methods with loop closure} \\
  \midrule
  \rowcolor{white} SLICT \cite{slict}               &            & \checkmark & 0.032     & 0.025     & 0.028     & \bf 0.023 & \bf 0.023 & 0.026     & \bf 0.030 & 0.029     & 0.034     & 0.028 \\
                   GLIM                             &            & \checkmark & \bf 0.025 & \bf 0.018 & \bf 0.023 & 0.025     & 0.027     & \bf 0.025 & 0.031     & \bf 0.024 & \bf 0.027 & \bf 0.025 \\
  \bottomrule
  \end{tabular}

  \vspace{2mm}
  {\bf Black} bold values are the best results for each configuration. \\
  {\dag} Both the horizontal and vertical LiDARs are used.
\end{table*}

\section{Discussion and Future Work}

Although we showed the robustness of the proposed framework to momentary degeneration of range data, it is still challenging for the proposed range-IMU odometry estimation to deal with long-term range data degeneration because we need to bound the duration of the optimization window of the odometry estimation for real-time performance. There are two possible approaches for making the system robust to long-term degeneration. The first approach is to incorporate an additional data source (e.g., a camera, radar, or wheel odometry) to suppress the estimation drift during the degeneration of one sensor. This approach has been successfully applied to visual-range-IMU methods \cite{Zheng_2022}. The second approach is to incorporate learning-based motion estimation into the odometry estimation. Recent pedestrian dead reckoning methods \cite{Herath_2020, Wang_2022} show that we can efficiently reduce trajectory estimation drift using only IMU data with a learning-based motion estimation model. These models enable the injection of prior knowledge of the motion pattern to reduce the uncertainty of the IMU-based velocity estimation during range data degeneration. We believe the extensibility of GLIM makes it easy to implement and evaluate both approaches for dealing with long-term range data degeneration.

Regarding global trajectory optimization, the experimental results show that the proposed GPU-accelerated global optimization algorithm can sufficiently handle practical situations. However, for more large-scale mapping problems (e.g., city- or nation-scale mapping), a more scalable optimization algorithm is needed. Distributed and GPU-suitable optimization algorithms such as Gaussian belief propagation \cite{futuremapping} are promising approaches for handling extremely large problems. The proposed GPU-based registration error factor would nicely fit with such GPU-based optimization algorithms.

\section{Conclusion}

This article presented GLIM, a 3D range-IMU SLAM framework with GPU acceleration. By fully leveraging the computation power of a modern GPU, we proposed odometry estimation and global trajectory optimization algorithms that enable robust and accurate localization and mapping in challenging situations. The experimental results show that state-of-the-art methods based on frame-to-model matching have difficulty dealing with completely degenerated range data whereas the proposed method can robustly estimate the sensor trajectory in such situations.

\balance

\bibliographystyle{IEEEtran}
\bibliography{ras2024}

\end{document}